\documentclass[lettersize,journal]{IEEEtran}
\usepackage{amsmath,amsfonts}
\usepackage{algorithmic}
\usepackage{algorithm}
\usepackage{makecell}
\usepackage{array}
\usepackage{textcomp}
\usepackage{stfloats}
\usepackage{url}
\usepackage{verbatim}
\usepackage{graphicx}
\usepackage{cite}

\usepackage{bm}
\usepackage{amssymb}
\usepackage{hyperref}
\usepackage{xcolor}
\usepackage{enumitem}
\usepackage{booktabs}
\usepackage{multirow}
\usepackage{orcidlink}
\usepackage{subfigure}
\usepackage{textcomp}
\usepackage[misc]{ifsym}
\usepackage[justification=centering]{caption}

\begin{document}

\title{Gaussian Process Latent Variable Modeling for Few-shot Time Series Forecasting}

\author{Yunyao Cheng~\orcidlink{0000-0002-1819-4056}, Chenjuan Guo~\textsuperscript{\Letter}~\orcidlink{0000-0002-4516-4637}, Kaixuan Chen~\orcidlink{0000-0003-3904-0395}, Kai Zhao~\orcidlink{0000-0002-5159-2312}, Bin Yang~\orcidlink{0000-0002-1658-1079}, \textit{Senior Member}, \textit{IEEE}, Jiandong Xie, Christian S. Jensen~\orcidlink{0000-0002-9697-7670}, \textit{Fellow}, \textit{IEEE}, Feiteng Huang, and Kai Zheng~\orcidlink{0000-0002-0217-3998}, \textit{Senior Member}, \textit{IEEE}

\IEEEcompsocitemizethanks{\IEEEcompsocthanksitem Y. Cheng, K. Chen, K. Zhao, and C. Jensen are with the Department
of Computer Science, Aalborg University, Denmark. E-mail: \{yunyaoc, kchen, kaiz, csj\}@cs.aau.dk
\IEEEcompsocthanksitem C. Guo, B. Yang are with the Department
of Computer Science, East China Normal University, China. E-mail: \{cjguo, byang\}@dase.ecnu.edu.cn
\IEEEcompsocthanksitem J. Xie and F. Huang are with Huawei Cloud Database Innovation Lab, China. E-mail: \{xiejiandong, huangfeiteng\}@huawei.com
\IEEEcompsocthanksitem K. Zheng is with the University of Electronic Science and Technology of China, China. E-mail: zhengkai@uestc.edu.cn}
} 


\markboth{Journal of \LaTeX\ Class Files,~Vol.~14, No.~8, August~2021}%
{Shell \MakeLowercase{\textit{et al.}}: A Sample Article Using IEEEtran.cls for IEEE Journals}


\maketitle

\begin{abstract}
Accurate time series forecasting is crucial for optimizing resource allocation, industrial production, and urban management, particularly with the growth of cyber-physical and IoT systems. However, limited training sample availability in fields like physics and biology poses significant challenges. Existing models struggle to capture long-term dependencies and to model diverse meta-knowledge explicitly in few-shot scenarios. To address these issues, we propose MetaGP, a meta-learning-based Gaussian process latent variable model that uses a Gaussian process kernel function to capture long-term dependencies and to maintain strong correlations in time series. We also introduce Kernel Association Search (KAS) as a novel meta-learning component to explicitly model meta-knowledge, thereby enhancing both interpretability and prediction accuracy. We study MetaGP on simulated and real-world few-shot datasets, showing that it is capable of state-of-the-art prediction accuracy. We also find that MetaGP can capture long-term dependencies and can model meta-knowledge, thereby providing valuable insights into complex time series patterns.
\end{abstract}


\section{Introduction} 
\label{introduction}

\IEEEPARstart{W}{ith} the ongoing digitalization of societal and industrial processes and the increasing deployment of cyber-physical and IoT systems, massive time series data are being generated to enable value creation through a variety of analyses. Accurate time series forecasting can improve resource allocation, optimize industrial production processes, and enhance urban management.
\begin{figure}[htbp]
\centering
\subfigure[Time series with both long-term and short-term seasonality]{
\includegraphics[width=0.23\textwidth]{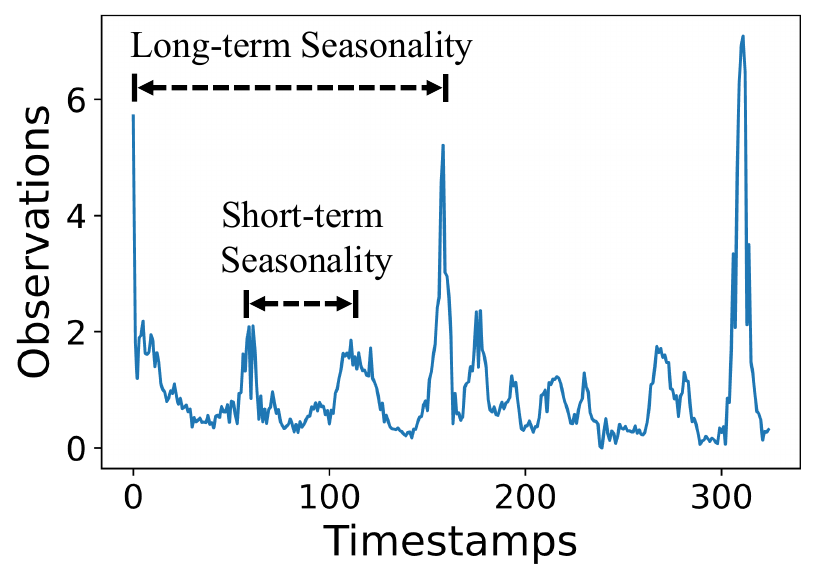}}
\subfigure[Time series with both trend and short-term seasonality]{
\includegraphics[width=0.23\textwidth]{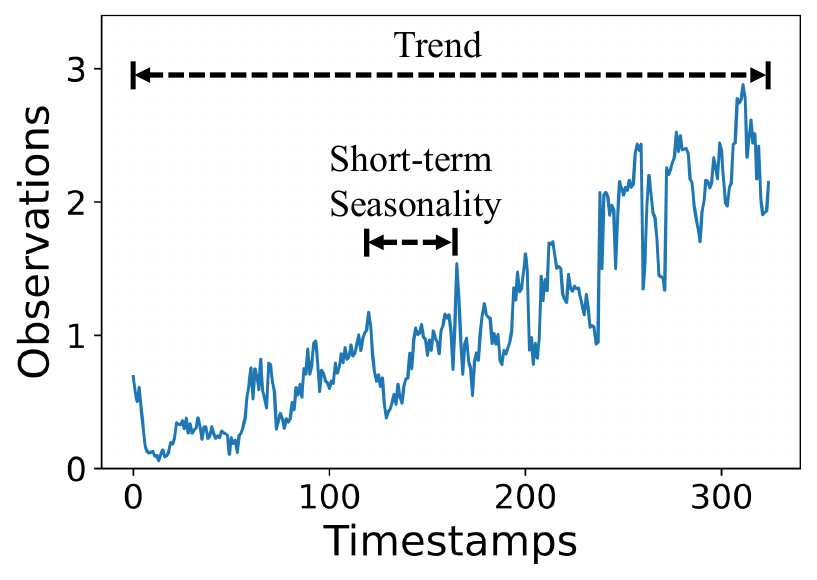}}
\caption{Time Series with Varying Patterns.}
\label{intro_ts}
\end{figure}

However, not all datasets possess sufficient numbers of training samples; this is particularly prevalent in physics and biology data. \autoref{intro_ts} shows two samples from a real-world biological dataset, depicting the changes in microbial community sizes over time. Due to high collection and analysis costs, the number of samples of microbial community data is limited. When a model is transferred to a new data distribution and the new data has insufficient samples, the model cannot be fully trained, which can cause overfitting and decrease prediction accuracy. Therefore, this paper investigates the problem of how to achieve time series forecasting models that can be applied to new data with only a few samples. We refer to this as few-shot time series forecasting.

\begin{figure}[htbp]
\centering
\subfigure[Sequential LVM]{
\includegraphics[width=0.23\textwidth]{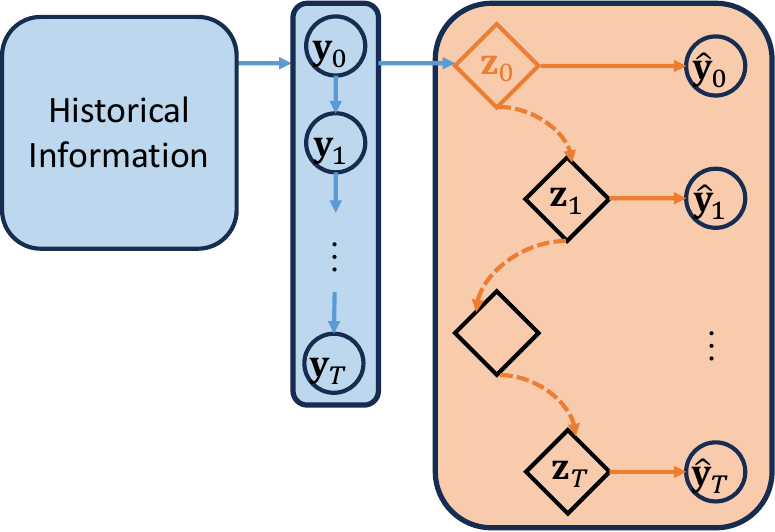}}
\subfigure[Meta-learning LVM]{
\includegraphics[width=0.23\textwidth]{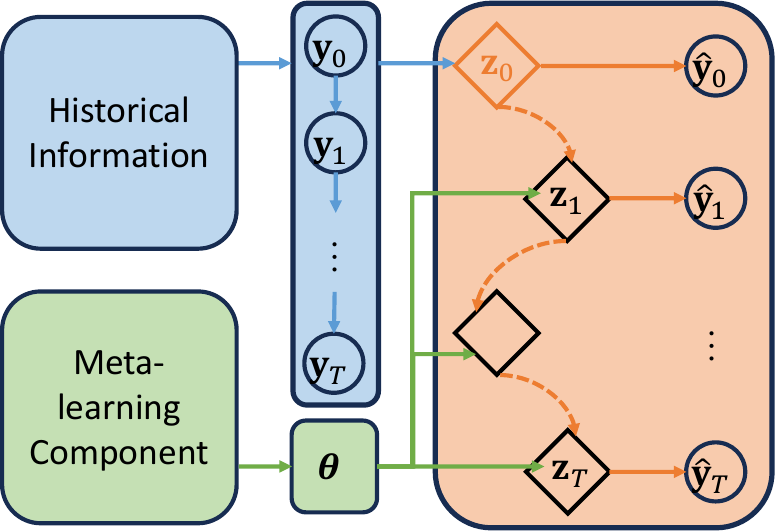}}
\caption{LVM Paradigms}
\label{intro_para}
\end{figure}

According to Bayesian theory, a well-performing model can be trained by enhancing likelihood (leveraging more data) or improving the prior (imposing constraints). Few-shot learning research focuses mainly on LLMs and hidden-state models: LLMs boost likelihood through large-scale pretraining, while hidden-state models refine priors with domain knowledge to improve generalization. However, LLMs struggle with scientific data due to their reliance on statistical learning over scientific laws, leading to poor generalization, violations of scientific principles, and difficulties with structured, multimodal, and low-resource datasets, exacerbated by scarce high-quality domain-specific data~\cite{sun2024scieval,zhong2024benchmarking}. Research on sequential latent variable models (LVMs)~\cite{jiang2022sequential,botev2021priors,fraccaro2017disentangled,karl2022deep} suggests that few-shot sequence forecasting can be improved by focusing on prior construction. Sequential LVMs can incorporate prior knowledge, for example, by making distributions of latent variables reflect domain-specific insights. Unlike autoregressive models, which establish correspondences between historical and future time series, sequential LVMs, as shown in~\autoref{intro_para}(a), encode historical information as an initial state $\mathbf{z}_{0}$ and obtain subsequent latent states through a latent dynamic function, $\mathbf{z}_{i}=f(\mathbf{z}_{i-1})$. A predicted observation is derived from an emission function $\hat{\mathbf{y}}_{i}=g(\mathbf{z}_{i})$. Therefore, sequential LVMs can infer the dynamics of an abstract latent state, not directly observed based on historical information, and can derive future observations. To enhance prediction accuracy in few-shot learning, as shown in~\autoref{intro_para}(b), Jiang et al.~\cite{jiang2022sequential} propose a meta-learning-based sequential LVM paradigm, where the latent dynamic function $\mathbf{z}_{i}=f(\mathbf{z}_{i-1};\bm{\theta})$ is fed meta-knowledge $\bm{\theta}$ as additional information. \textbf{Patterns such as trends and seasonality in time series, as shown in~\autoref{intro_ts}, which are presented in an explicit and quantifiable form, can be regarded as meta-knowledge.} The meta-knowledge $\bm{\theta}$ is generated by the meta-learning component that can be a trainable neural network.

Although existing methods for few-shot sequence forecasting achieve significant advancements, two important challenges remain to be addressed.

\noindent {\bf Challenge 1:} It is challenging for existing few-shot learning methods to achieve high forecasting accuracy on time series with \textbf{long-term dependencies}. These dependencies refer to patterns such as trend and seasonality that span an entire time series, and they are prevalent in time series. In contrast, existing sequential LVMs~\cite{jiang2022sequential,botev2021priors,fraccaro2017disentangled,karl2022deep} focus mainly on image sequence forecasting, where they predict future frames based on the current frame and where short-term dependencies occur due to the continuity of object motion. Thus, LVMs typically rely on short-term modeling, where the current latent state is derived from the previous one. Although attention mechanisms~\cite{nie2022time,DBLP:conf/ijcai/CirsteaG0KDP22} help capture global dependencies, their performance deteriorates in few-shot settings due to high model complexity, which increases the risk of overfitting. Moreover, these methods lack explicit meta-learning capabilities and must relearn attention weights from scratch for each new task using limited data, which hinders generalization. As a result, capturing long-term dependencies remains challenging.

\noindent {\bf Challenge 2:} It is challenging for existing methods to explicitly model and capture \textbf{varying meta-knowledge}. As shown in~\autoref{intro_ts}, time series often exhibit varying meta-knowledge, making it difficult for a single dynamic function to capture diverse latent dynamics. The meta-learning component proposed by Jiang et al.~\cite{jiang2022sequential} is a black-box neural network, resulting in a simple and implicit, rather than complex and explicit, learning of meta-knowledge. This makes it difficult for the model to capture complex patterns. Additionally, explicit meta-knowledge is crucial in many analyses, as it provides users with information on seasonality and other relevant features.

To address these challenges, we propose MetaGP, a meta-learning-based Gaussian process latent variable model. MetaGP tackles the two challenges as follows.

\noindent {\bf Addressing challenge 1:} To enable MetaGP to capture long-term dependencies in time series in few-shot learning, we propose a novel Gaussian process latent variable model based on LVM and meta-learning theories. A Gaussian process~\cite{jankowiak2020deep,salimbeni2017doubly} models stochastic functions, with its kernel function defining the measure of similarity in the input space. The kernel function can ensure that observations far apart still maintain strong correlations, thereby enabling the capture of long-term dependencies. Hence, we derive and design a Gaussian process as the dynamic function of the LVM, with its kernel function serving as the meta-knowledge for meta-learning, to address the challenge of capturing long-term dependencies in time series in the few-shot learning, which is difficult for sequential LVMs.

\noindent {\bf Addressing challenge 2:} To enable MetaGP to explicitly model and capture varying meta-knowledge, we propose a novel meta-learning component called Kernel Association Search (KAS) by neural architecture search and the kernel association theories~\cite{wu2021autocts,duvenaud2013structure}. Unlike traditional meta-learning components with black-box meta-knowledge, MetaGP adopts an explicit approach to generate meta-knowledge. The meta-knowledge in MetaGP is the kernel function of the Gaussian process LVM, where kernel association theory ensures that more complex kernel functions can be generated from a simple prior which is a collection of basic kernel functions. Both the prior and the final learned meta-knowledge possess analytic forms, making them explicit. Then, to enhance KAS to handle more complex patterns in multivariate time series, we propose cross-variable weights, which is a learnable parameter matrix used to control the weights of correlations among variables. Furthermore, KAS utilizes neural architecture search theory to provide an end-to-end meta-learning approach for MetaGP, enabling flexible integration with neural networks and Gaussian process LVMs.

In summary, it is crucial to enable LVMs to capture long-term dependencies in few-shot time series forecasting. Moreover, modeling and capturing varying meta-knowledge explicitly is essential to improve prediction accuracy and interpretability. We make three main contributions:
\begin{itemize}[leftmargin=*]
\item We propose a novel meta-learning Gaussian process latent variable model (MetaGP) that, in few-shot learning, can capture long-term dependencies in time series.
\item We propose KAS, a novel meta-learning component that can model and capture varying meta-knowledge and correlations in multivariate time series explicitly.
\item We report on extensive experiments on simulated physical and real-world biological few-shot datasets, finding that MetaGP is capable of state-of-the-art prediction accuracy. We validate the advantages of MetaGP at capturing long-term dependencies in time series. Furthermore, we analyze and visualize the varying meta-knowledge and correlations captured by KAS.
\end{itemize}

\section{Related Work}
\label{relatedwork}
\noindent{\bf Few-shot Learning for Time Series:} Few-shot learning is commonly applied to classification and forecasting tasks on time series data. Zhong et al.~\cite{zhong2023online} employ meta-learning approaches, integrating time series classification models with domain knowledge from the geological sector to detect aftershocks. Chen et al.~\cite{chen2023supervised} utilize contrastive learning to alleviate the few-shot challenge in high-frequency time series classification. Montero-Manso et al.~\cite{montero2020fforma} and Oreshkin et al.~\cite{oreshkin2021meta} report the first attempts to address few-shot learning for univariate time series forecasting. Montero-Manso et al.~\cite{montero2020fforma} employ statistical methods to weight and combine forecast results, while Oreshkin et al.~\cite{oreshkin2021meta} combine meta-learning with neural networks, both incorporating strategies that leverage limited samples and temporal domain knowledge. Jiang et al.~\cite{jiang2022sequential} introduce a sequential latent variable model that uses the support set as a prior, combined with knowledge derived from few-shot support series, to mitigate the few-shot challenge in deriving dynamic functions. In summary, existing approaches to few-shot learning in scientific time series data predominantly adopt Bayesian principles and utilize small amounts of data and time series-related domain knowledge as priors to address this challenge. However, these methods face challenges in handling long-term dependencies and capturing meta-knowledge.

\noindent{\bf Long-term Dependencies Learning:} Existing approaches to long-term dependency learning in time series forecasting rely mainly on attention mechanisms. The Transformer~\cite{vaswani2017attention} is adopted due to its strength in modeling long-range dependencies via self-attention. LogTrans~\cite{li2019enhancing} introduces sparse attention for this purpose. Subsequent methods, such as Informer~\cite{zhou2021informer}, Autoformer~\cite{wu2021autoformer}, Triformer~\cite{DBLP:conf/ijcai/CirsteaG0KDP22}, Pyraformer~\cite{liu2021pyraformer}, and FEDformer~\cite{zhou2022fedformer}, propose attention variants tailored to temporal dynamics. Transformer-based models have since evolved in two directions: channel-independent approaches (e.g., PatchTST~\cite{nie2022time} and Pathformer~\cite{chen2024pathformer}), focusing on intra-channel patterns; and cross-channel models (e.g., Crossformer~\cite{zhang2022crossformer}, CARD~\cite{wang2023card}, and iTransformer~\cite{liu2023itransformer}), focusing on the explicit capture of inter-channel dependencies to enhance long-term forecasting performance. However, these methods face challenges in few-shot learning scenarios. Attention mechanisms tend to overfit when training samples are limited, and they lack meta-learning capabilities that can enable them to generalize effectively under data scarcity.

\noindent {\bf Gaussian Processes:} 
Applications of Gaussian processes in neural networks occur in two settings---deep Gaussian processes and deep kernel learning. Deep Gaussian processes~\cite{jankowiak2020deep,salimbeni2017doubly,damianou2013deep} replace each layer of a neural network with a Gaussian process to overcome overfitting. However, it is difficult to study correlations of time series explicitly with these methods. Next, deep kernel learning combines the structural properties of deep architectures with the non-parametric flexibility of kernel methods. Wilson et al.~\cite{wilson2016deep,wilson2016stochastic} combine a neural network and a Gaussian process layer with a spectral mixture kernel. Al-Shedivat et al.~\cite{al2017learning} propose to learn kernels with LSTMs so that the kernels encapsulate the properties of LSTMs. However, these methods lack the ability to model the diversity seen in multivariate time series. In addition, to capture the multivariate covariance among time series, multi-task Gaussian process models~\cite{bonilla2007multi,wilson2012gaussian,titsias2011spike,guarnizo2015indian} utilize fixed and pre-defined kernel functions, which limit their ability to capture diverse correlations.

\noindent {\bf Kernel Learning:} 
Gaussian process-based kernel learning methods are used for time series forecasting. These methods generate new kernels by combining existing kernels~\cite{schulz2017compositional} via kernel association. A selection strategy determines how to select the best kernels, using, e.g., greedy search~\cite{duvenaud2013structure,kim2018scaling,lu2018structured,schulz2017compositional,tong2019discovering}. However, existing kernels learning approaches suffer from high time complexity. These methods utilize the Bayesian information criterion to select optimal kernel functions. Sun et al.~\cite{sun2018differentiable} propose a neural network that alternates between stacked linear and interactive layers to simulate kernel association. However, the capacity of the kernel function space is fixed, which may lead to sub-optimal kernels. We propose a dynamic and efficient kernel learning framework that is integrated well into a deep architecture.
\section{Preliminaries} 
\label{prelinimaries}
\begin{figure*}[htbp]
\centering
\subfigure[Classical Gaussian Process]{
\label{fig:subfig:GP} 
\includegraphics[width=0.35\textwidth]{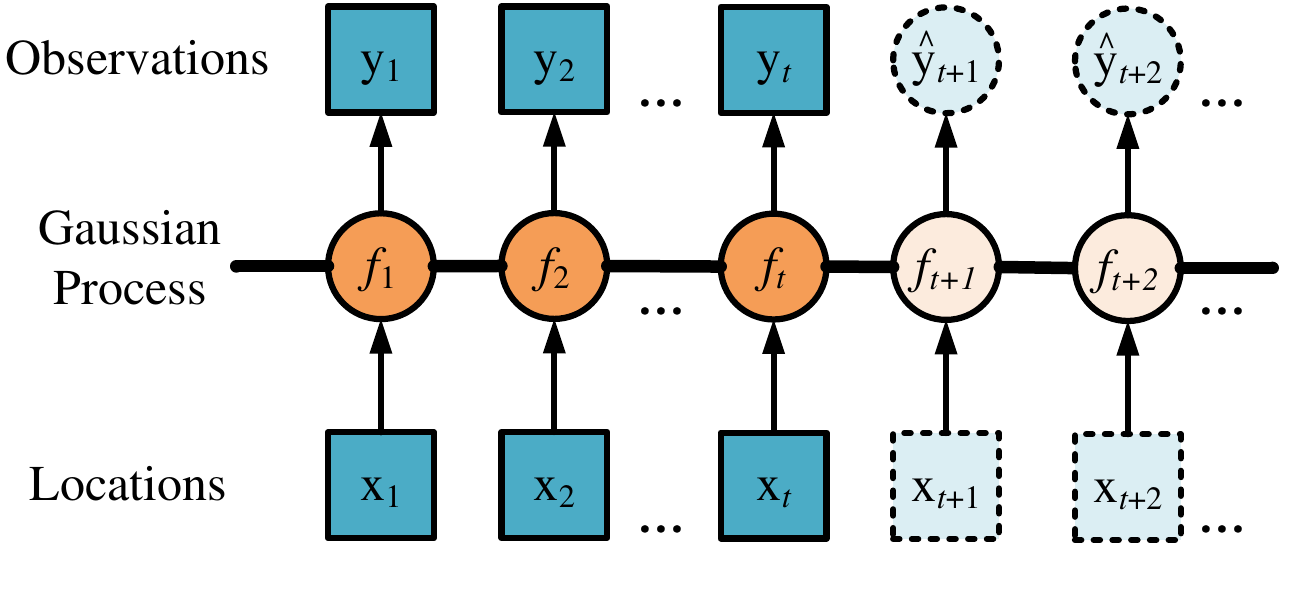}}
\subfigure[Gaussian Process Latent Variable Model]{
\label{fig:subfig:Auto GP} 
\includegraphics[width=0.55\textwidth]{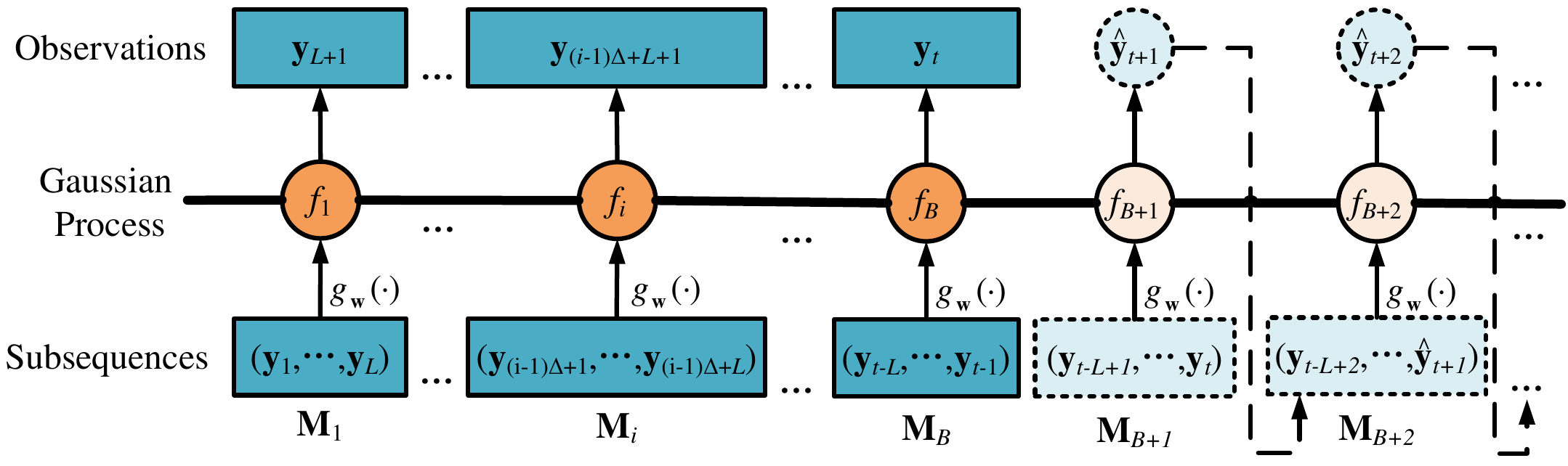}}
\caption{Gaussian Processes.}
\label{fig:GPS} 
\end{figure*}

\subsection{Problem Definition}
We consider a multivariate time series $\mathbf{Y}\in \mathbb{R}^{t\times N}$, where $t$ denotes the number of observations and $N$ is the number of time series, or variables. An $F$-timestamps-ahead time series forecasting model $\mathcal{F}$ takes as input $L$ historical observations $\mathbf{y}_{t-L+1},\mathbf{y}_{t-L+2},...,\mathbf{y}_{t}$, with $L\ll t$, and predicts the $F$ future observations $\mathbf{y}_{t+1},\mathbf{y}_{t+2},...,\mathbf{y}_{t+F}$:
\begin{equation}
\setlength{\abovedisplayskip}{3pt}
\setlength{\belowdisplayskip}{3pt}
    \mathcal{F}_{\bm{\Gamma}}(\mathbf{y}_{t-L+1},\mathbf{y}_{t-L+2},...,\mathbf{y}_{t}) =
   (\mathbf{y}_{t+1},\mathbf{y}_{t+2},...,\mathbf{y}_{t+F}),
\end{equation}
where $\Gamma$ is a collection of learnable parameters and functions.

\subsection{Gaussian Process}\label{Gaussian Process}

A Gaussian process is a joint Gaussian distribution consisting of a set of random variables~\cite{williams2006gaussian}. Since training and inference proceed differently, we introduce them separately. 

\noindent\textbf{Training:} As Fig.~\ref{fig:GPS}(a) shows, given $t$ training samples of location and observation pairs $(\mathbf{x},\mathbf{y})=\{(\mathrm{x}_i,\mathrm{y}_i)|i=1,2,...,t\}$ with additive, independent, and identically distributed Gaussian noise $\varepsilon$, we have $\mathrm{y}_i=f(\mathrm{x}_i)\ +\ \varepsilon$. For simplicity, we use $f(\mathrm{x}_i)$ and $f_i$ interchangeably, and we assume that $\mathbf{f}=f(\mathbf{x})=(f_1,f_2,...,f_i,...,f_t)$ is a collection of random variables. We assume a traditional Gaussian process and use ``locations" in place of timestamps. A Gaussian process can be specified by its given priors---its mean $\mu(\mathrm{x}_i)$ and kernel function $k_{\theta}(\mathrm{x}_i,\mathrm{x}_j)$ that is parameterized by $\theta$. Specifically, $\theta$ can be given as prior values or learned through optimization~\cite{wilson2014covariance}. The kernel function associates any pair of two random variables at locations $\mathrm{x}_i$ and $\mathrm{x}_j$. We use $i$, $j$ to denote any two indices in a subsequence or a kernel. The mean and kernel functions are denoted as:
\begin{equation}
\begin{aligned}
    \mu(\mathrm{x}_i) &= \mathbb{E}[f(\mathrm{x}_i)] \\
    k_{\theta}(\mathrm{x}_i,\mathrm{x}_j) &= \mathbb{E}[(f(\mathrm{x}_i)-\mu(\mathrm{x}_i))(f(\mathrm{x}_j)-\mu(\mathrm{x}_j))],
\end{aligned}
\end{equation}

\noindent giving rise to the Gaussian process:
\begin{equation}
\begin{aligned}
    \mathbf{f} \sim \mathcal{GP}(\bm{\mu}(\mathbf{x}),\mathbf{K}(\mathbf{x},\mathbf{x})), 
\end{aligned}
\end{equation}
where $\bm{\mu}(\mathbf{x}) = (\mu(\mathrm{x}_1), \mu(\mathrm{x}_2),..., \mu(\mathrm{x}_t))$ denotes the mean vector containing values for all training observations and $\mathbf{K}(\mathrm{x}, \mathrm{x})$ is the covariance matrix that contains covariance values computed by kernel function $k_{\theta}(\mathrm{x}_i, \mathrm{x}_j)$ for each pair of training locations $\mathrm{x}_{i}$ and $\mathrm{x}_{j}$.

Although some studies refer to kernel functions and covariance matrices interchangeably, we distinguish between them. Specifically, we use kernel functions to represent the analytic forms of kernels that define priors on a Gaussian process, and we use covariance matrices to represent the matrices of the covariance values computed by kernel functions.

\noindent\textbf{Inference:} Inference aims to predict random variables $\mathbf{f}_*=(f_{t+1},f_{t+2},...)$ at test locations $\mathbf{x}_*=(\mathrm{x}_{t+1},\mathrm{x}_{t+2},...)$. Assuming noise $\varepsilon$ with variance $\sigma^2$, we formulate the distribution of the training observations $\mathbf{y}$ and the prediction random variables $\mathbf{f}_*$ under the priors, i.e., kernel functions, as follows.
\begin{equation}\label{function (3)}
\begin{aligned}
    \begin{bmatrix}
     \mathbf{y}\\
     \mathbf{f}_*
    \end{bmatrix}
    =
    \mathcal{N} \left(
    \begin{bmatrix}
     \bm{\mu}(\mathbf{x})\\
     \bm{\mu}(\mathbf{x}_*)
    \end{bmatrix}
    ,
    \begin{bmatrix}
     \mathbf{K}(\mathbf{x},\mathbf{x})+\sigma^2\mathbf{I} & \mathbf{K}(\mathbf{x},\mathbf{x}_*)\\
     \mathbf{K}(\mathbf{x}_*,\mathbf{x}) & \mathbf{K}(\mathbf{x}_*,\mathbf{x}_*)
    \end{bmatrix} \right),
\end{aligned}
\end{equation}
where $\bm{\mu}(\mathbf{x})$ and $\bm{\mu}(\mathbf{x}_*)$ denote the mean vectors and $\mathbf{K}(\mathbf{x},\mathbf{x})$, $\mathbf{K}(\mathbf{x}_*,\mathbf{x})$, $\mathbf{K}(\mathbf{x},\mathbf{x}_*)$, and $\mathbf{K}(\mathbf{x}_*,\mathbf{x}_*)$ are the covariance matrices. Then the distribution of the prediction $\mathbf{f}_*$ has the form:
\begin{equation}\label{function (4)}
\begin{aligned}
    \mathbf{f}_*|\mathbf{x},\mathbf{y},\mathbf{x}_* \sim \mathcal{N}(\mathbb{E}[\mathbf{f}_*],\mathrm{Cov}[\mathbf{f}_*]),
\end{aligned}
\end{equation}
where
\begin{equation}
\begin{aligned}
    \mathbb{E}[\mathbf{f}_*] &= \bm{\mu}(\mathbf{x}_*) + \mathbf{K}(\mathbf{x}_*,\mathbf{x})[\mathbf{K}(\mathbf{x},\mathbf{x})+{\sigma}^2\mathbf{I}]^{-1}(\mathbf{y}-\bm{\mu}(\mathbf{x}))\label{con:GP_inference} \\
    \mathrm{Cov}[\mathbf{f}_*] &= \mathbf{K}(\mathbf{x}_*,\mathbf{x}_*) - \mathbf{K}(\mathbf{x}_*,\mathbf{x})[\mathbf{K}(\mathbf{x},\mathbf{x})+{\sigma}^2\mathbf{I}]^{-1}\mathbf{K}(\mathbf{x},\mathbf{x}_*)
\end{aligned}
\end{equation}

When we conduct point estimation, i.e., predict $\mathbf{y}_*$ given $\mathbf{x}_*$, we do not use the random variables $\mathbf{f}_*$ as the prediction values. Instead, we use the expectations $\mathbb{E}[\mathbf{f}_*]$ as the predictions, i.e., $\hat{\mathbf{y}}_*=(\hat{\mathrm{y}}_{t+1},\hat{\mathrm{y}}_{t+2},...)=(\mathbb{E}(f_{t+1}),\mathbb{E}(f_{t+2}),...)$, and the standards deviations $\mathrm{Cov}[\mathbf{f}_*]$ can be used to obtain confidence intervals. 

\subsection{Kernel Functions}
Kernel functions are the priors used by a Gaussian process and are the analytic forms of the functions that we are modeling. A kernel function is a function of $\mathrm{x}_i-\mathrm{x}_j$~\cite{williams2006gaussian}. A kernel function thus specifies the covariance between a pair of random variables using a function of their location distance. For example, the squared exponential (SE) kernel is one of the most commonly used kernel functions in Gaussian processes. The observations $\mathrm{y}_i$ and $\mathrm{y}_j$ at two nearby locations $\mathrm{x}_i$ and $\mathrm{x}_j$ are expected to be close in a Gaussian process. In other words, the closer the locations, the more similar the observations should be. Thus, the SE kernel function represents an interpretable pattern that captures short-term dependencies. 

The SE kernel is defined as:
\begin{equation}
\begin{aligned}
    k_{\theta}(\mathrm{x}_i,\mathrm{x}_j) = \sigma_k^2\mathrm{exp}\left(-\frac{||\mathrm{x}_i-\mathrm{x}_j||}{2l^2}\right),
\end{aligned}
\end{equation}
where $\theta=\{\sigma_k,l\}$ denotes the parameters. Specifically, $\sigma_k$ denotes the scale factor of the covariance and $l$ represents the parameter length-scale. A shorter length-scale indicates that observations $\mathbf{y}$ vary more rapidly across locations $\mathbf{x}$.
\begin{table}[htbp]
\setlength\tabcolsep{7pt}
\footnotesize
\centering
\caption{Basic Kernel Functions.}
\label{Base_kernel_functions}
\scalebox{0.75}{
\begin{tabular}{|c|c|c|c|}
\hline
\begin{tabular}[c]{@{}c@{}}Kernel Function\end{tabular} & \begin{tabular}[c]{@{}c@{}}Interpretable Pattern\end{tabular} & Analytic Form                                                                                               & Parameter(s) \\ \hline
SE                                                         & Short-term                                                      & $\sigma_k^2\mathrm{exp}\left(-\frac{||\mathrm{x}_i-\mathrm{x}_j||}{2l^2}\right)$                            & $l$          \\
PER                                                        & Seasonality                                                     & $\sigma_k^2\mathrm{exp}\left(-\frac{2\mathrm{sin}^2(\frac{\pi|\mathrm{x}_i-\mathrm{x}_j|)}{p}}{l^2}\right)$ & $l$,$p$      \\
LIN                                                        & Trend                                                           & $\sigma_k^2(\mathrm{x}_i-c)(\mathrm{x}_j-c)$                                                                & $c$          \\
RQ                                                         & Long-term                                                       & $\sigma_k^2\left(1+\frac{||\mathrm{x}_i-\mathrm{x}_j||}{2al^2}\right)^{-a}$                     & $l$,$a$ $>0$     \\ \hline
\end{tabular}
}
\end{table}

Numerous kernel functions exist beyond the SE kernel function that describe observations in an interpretable way.~\autoref{Base_kernel_functions} lists basic kernel functions and the patterns they target. The periodic analytic form of the PER kernel captures periodic correlations, while the linear analytic form of the LIN kernel captures linear correlations. The RQ kernel reduces to the SE kernel when its parameter $a \rightarrow \infty$~\cite{rasmussen2003gaussian}. The covariance captured via the RQ kernel decays more slowly with increasing distance than in the case of the SE kernel, as SE is the inverse of an exponential function and RQ is a power function with power $-a$. Therefore, the RQ kernel can capture long-term dependencies. Another study~\cite{duvenaud-thesis-2014} provides a detailed explanation of the mathematical properties of different kernel functions, thus serving as a comprehensive kernel guide.
\newtheorem{theorem}{Theorem}[section]
\newtheorem{corollary}{Corollary}[theorem]
\newtheorem{lemma}[theorem]{Lemma}

\section{Methodology} 
\label{methodology}

\subsection{MetaGP Framework}
\label{MetaGP}
\begin{figure}[htbp]
  \centering
  \includegraphics[width=\linewidth]{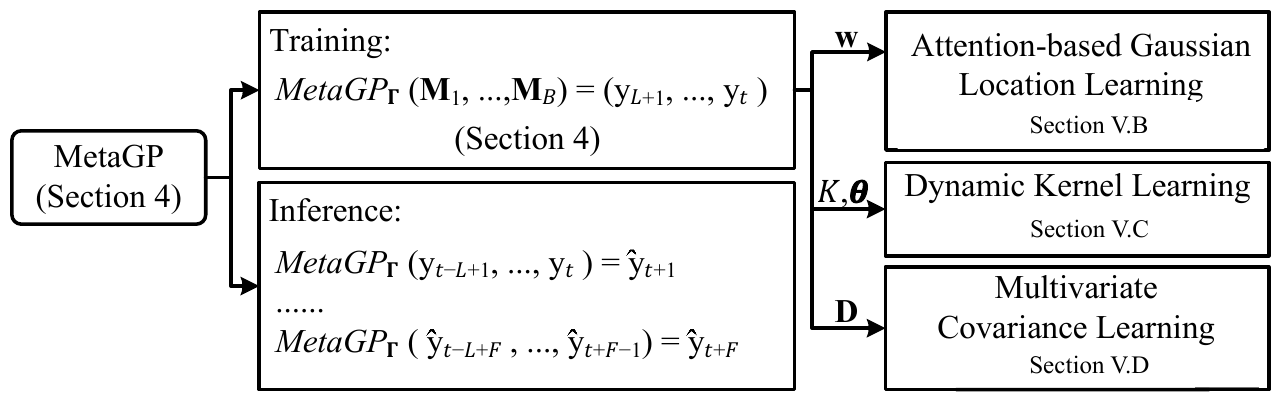}
  \caption{MetaGP Framework Encompassing Training and Inference.}
  \label{fig:Overview}
\end{figure}

Figure~\ref{fig:Overview} illustrates the overall framework of MetaGP, which covers two main stages: training and inference. In the training stage, the model learns a Gaussian process-based dynamic function by processing historical time series segments. These segments are used to predict the next-step observations, enabling the model to capture temporal patterns and dependencies.

The learned model consists of three core components. First, an attention-based module learns the temporal structure and highlights important time steps. Second, a dynamic kernel learning module builds flexible kernel functions to model complex dependencies. Third, a multivariate covariance learning component captures correlations across different variables in the time series.

In the inference stage, the model generates future predictions iteratively. It takes recent observations (or previously predicted values) as input and forecasts the next time step. This step-by-step process allows the model to incorporate newly predicted information adaptively and to produce a sequence of future values. Together, these modules aim to enable MetaGP to perform accurate and interpretable few-shot time series forecasting, even in complex and data-scarce scenarios.

Traditional Gaussian process models use timestamps as the Gaussian locations for regression tasks. In other words, they predict observations given their locations or timestamps. In contrast, Gaussian process LVMs generate latent variables sequentially using historical observations in time order. We define a Gaussian process LVM for multivariate time series forecasting that recursively predicts each future observation given a subsequence of historical observations. Further, the latent variables in the full time series are modeled using Gaussian process kernel functions. More specifically, this process contains two procedures---training and inference.

\noindent {\bf Training:} In order to establish a Gaussian process dynamic function of latent variables as exemplified in Fig.~\ref{fig:GPS}(b), the multivariate time series $\mathbf{Y}\in \mathbb{R}^{t\times N}$ is sliced into a series $\mathbf{M}$ of $B$ subsequences using moving a time window of length $L$ and step size $\Delta$: $\mathbf{M}=(\mathbf{M}_1,\mathbf{M}_2,...,\mathbf{M}_i,...,\mathbf{M}_B)$, where $\mathbf{M}_i\in \mathbb{R}^{L\times N}$ and $B=(t-L+1)/\Delta$. The $i$-th subsequence $\mathbf{M}_i$ is denoted as:
\begin{equation}
\begin{aligned}
    \mathbf{M}_i = (\mathbf{y}_{(i-1)\Delta+1},\mathbf{y}_{(i-1)\Delta+2},...,\mathbf{y}_{(i-1)\Delta+L})
\end{aligned}
\end{equation}

As in traditional Gaussian process training, we define the expectation of a future observation as $\mathbf{y}_{(i-1)\Delta+L+1}=\mathbb{E}(f_i)$, where $f_i=f(g_{\mathbf{w}}(\mathbf{M}_i))$ and $g_{\mathbf{w}}(\cdot)$ is a neural network used to encode the historical observations, as shown in Fig.~\ref{fig:GPS}(b). To enable time series forecasting, we build a temporal correlation between a historical subsequence $\mathbf{M}_i$ and a predicted observation $\mathbf{y}_{(i-1)\Delta+L+1}$. We then match $g_{\mathbf{w}}(\mathbf{M}_i)$ and $\mathbf{y}_{(i-1)\Delta+L+1}$ with $\mathrm{x}_i$ and $\mathrm{y}_i$ in the training stages as shown in Sec.~\ref{Gaussian Process}, and then we formulate the dynamic function as follows.
\begin{equation}
\begin{aligned}
   (f_1,...,f_i,...,f_B) \sim \mathcal{GP}(\bm{\mu}(g_{\mathbf{w}}(\mathbf{M})),\mathbf{K}(g_{\mathbf{w}}(\mathbf{M}),g_{\mathbf{w}}(\mathbf{M}))),
\end{aligned}
\end{equation}
where $\bm{\mu}(g_{\mathbf{w}}(\mathbf{M}))=(\mu(g_{\mathbf{w}}(\mathbf{M}_1)),...,\mu(g_{\mathbf{w}}(\mathbf{M}_B)))$ is the mean vector and $\mathbf{K}(g_{\mathbf{w}}(\mathbf{M}),g_{\mathbf{w}}(\mathbf{M}))$ is the covariance matrix. 

To achieve better forecasting accuracy, our goal is to learn the MetaGP model which contains three parts. The first is a neural network $g_{\mathbf{w}}(\cdot)$ that encodes the subsequences $\mathbf{M}$ as latent variables. The second is a final kernel function $K_{\bm{\theta}}$ that is an association of kernel functions guided by a learnable cross-variable weight matrix $\mathbf{D}$, which is the third part. Here $g_{\mathbf{w}}(\cdot)$ is parameterized by $\mathbf{w}$, and $K_{\bm{\theta}}$ is parameterized by $\bm{\theta}$. Thus, the model has parameters $\bm\Gamma=\{\mathbf{w},K,\bm{\theta}, \mathbf{D}\}$ and can be stated as follows.
\begin{equation}
\begin{aligned}
    \mathit{MetaGP}_{\bm{\Gamma}}(\mathbf{M}_1,...,\mathbf{M}_B)&=(\mathbf{y}_{L+1},...,\mathbf{y}_t)
\end{aligned}
\end{equation}

We train the model using $\mathbf{M}_1,...,\mathbf{M}_B$, and $\mathbf{y}_{L+1},...,\mathbf{y}_t$, in order to learn ${\mathbf{w}}$, $K$, ${\bm{\theta}}$, and $\mathbf{D}$.

\noindent {\bf Inference:} To iteratively infer $F$ future observations, as illustrated in Sec.~\ref{Gaussian Process}, we use Gaussian process inference. Specifically, in the first iteration, we match $g_{\mathbf{w}}(\mathbf{M}_{B+1})$, where $\mathbf{M}_{B+1}=\mathbf{y}_{t-L+1},...,\mathbf{y}_t$, and $\hat{\mathbf{y}}_{t+1}$ with $\mathrm{x}_{t+1}$ and $\hat{\mathrm{y}}_{t+1}$, and thus infer $\mathbf{y}_{t+1}$ using Equations~\ref{function (3)}--6, as shown in Fig.~\ref{fig:GPS}(b). Unlike Gaussian process inference, which takes arbitrary future locations as input and predicts their observations, Gaussian process LVMs predict iteratively, as follows.
\begin{equation}\label{MetaGP inference}
\begin{aligned}
   \mathit{MetaGP}_{\bm{\Gamma}}(\mathbf{y}_{t-L+1},...,\mathbf{y}_{t}) &= \mathbf{\hat{y}}_{t+1} \\
   \mathit{MetaGP}_{\bm{\Gamma}}(\mathbf{y}_{t-L+2},...,\mathbf{\hat{y}}_{t+1}) &= \mathbf{\hat{y}}_{t+2} \\
   ... \\
   \mathit{MetaGP}_{\bm{\Gamma}}(\mathbf{\hat{y}}_{t-L+F},...,\mathbf{\hat{y}}_{t+F-1}) &= \mathbf{\hat{y}}_{t+F},
\end{aligned}
\end{equation}
where the sequence $\mathbf{\hat{y}}_{t+1},\mathbf{\hat{y}}_{t+2},...,\mathbf{\hat{y}}_{t+F}$ denotes $F$ predictions.

\subsection{Attention-based Gaussian Location Learning}\label{attention-based Gaussian locations learning}
To capture and highlight salient temporal features in each subsequence, we propose an attention-based Gaussian location learning mechanism. Unlike traditional methods that treat all time steps equally, the attention mechanism dynamically assigns different weights to the input, enabling selective extraction of informative temporal patterns. This enhances the quality of latent representations with the goal of improving the accuracy and interpretability of MetaGP.

\begin{figure*}[htbp]
  \centering
  \includegraphics[width=0.9\linewidth]{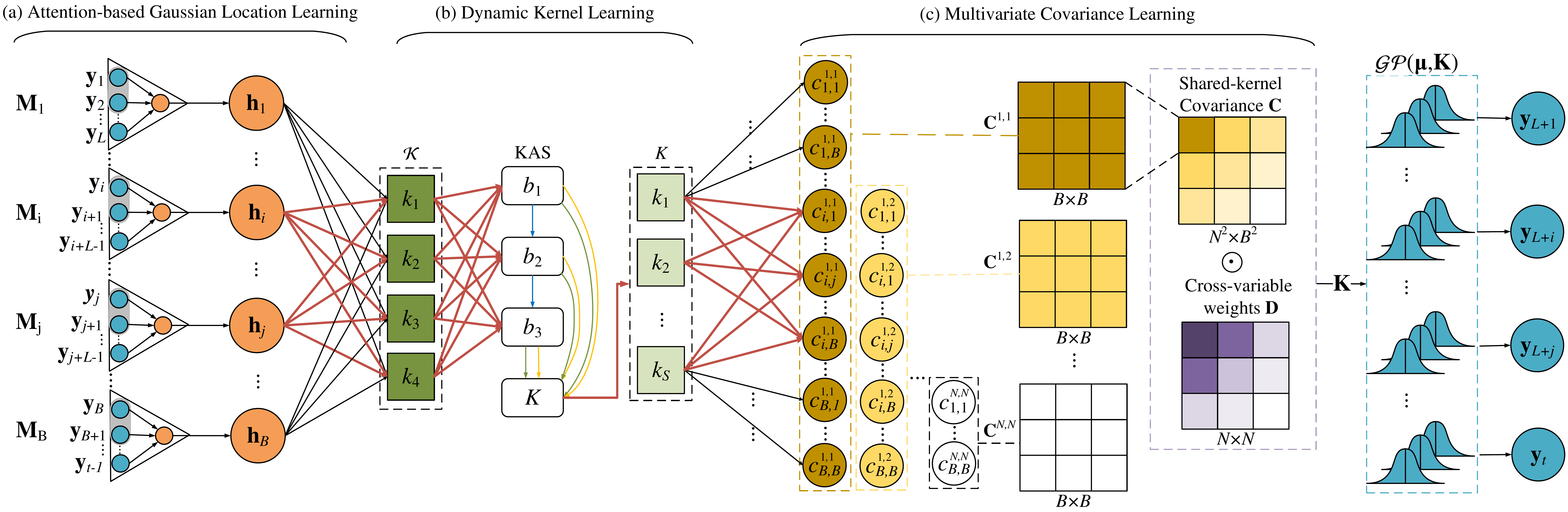}
  \caption{MetaGP Training.}
  \label{fig:DMDKL}
\end{figure*}

The training of the MetaGP framework is illustrated in Fig.~\ref{fig:DMDKL}. The framework takes as input subsequences $\mathbf{M}=(\mathbf{M}_1,\mathbf{M}_2,...,\mathbf{M}_i,...,\mathbf{M}_B)$, where $\mathbf{M}_i\in \mathbb{R}^{L\times N}$, and outputs predictions $\mathbf{y}_{L+1},...,\mathbf{y}_{L+i},...,\mathbf{y}_{L+j},...,\mathbf{y}_t$ for training.

The first component, attention-based Gaussian location learning, takes $\mathbf{M}$ as input and outputs latent variables $\mathbf{H}=(\mathbf{h}_1,...,\mathbf{h}_i,...,\mathbf{h}_j,...,\mathbf{h}_B)$, where $\mathbf{h}_i \in \mathbb{R}^{N}$ represents the location of a Gaussian process; see Fig.~\ref{fig:DMDKL}(a). We use patch-attention~\cite{DBLP:conf/ijcai/CirsteaG0KDP22} to capture the temporal dynamics in subsequence $\mathbf{M}_i$ and encode subsequences $\mathbf{M}$ to Gaussian locations $\mathbf{H}$. MetaGP utilizes attention structures to capture local features in subsequence $\mathbf{M}_i$ and utilizes the Gaussian process to construct global features among all subsequences in $\mathbf{M}$. This enables MetaGP to capture multiscale information from time series, and short-term and long-term dependencies.

The mechanism to achieve the attention-based Gaussian locations $\mathbf{H}$ is defined as follows.
\begin{equation}
\begin{aligned}
    \mathbf{H} &= (\mathbf{h}_1,...,\mathbf{h}_i,...,\mathbf{h}_j,...,\mathbf{h}_B)\\
               &= (g_{\mathbf{w}}(\mathbf{M_1}),...,g_{\mathbf{w}}(\mathbf{M_i}),...,g_{\mathbf{w}}(\mathbf{M_j}),...,g_{\mathbf{w}}(\mathbf{M_B}))
\end{aligned}
\end{equation}

Next, representations $\mathbf{H}$ are input into the Gaussian layer, which is formulated as follows.
\begin{equation}
\begin{aligned}
    (f_1,...,f_i,...,f_j,...,f_B) \sim \mathcal{GP}(\bm{\mu}(\mathbf{H}),\mathbf{K}(\mathbf{H},\mathbf{H})),
\end{aligned}
\end{equation}
where $\bm{\mu}(\mathbf{H})=(\mu(\mathbf{h_1}),...,\mu(\mathbf{h_i}),...,\mu(\mathbf{h_j}),...,\mu(\mathbf{h_B}))$ and $\mathbf{K}(\mathbf{H},\mathbf{H})$ are the mean vector and the covariance matrix.

\subsection{Dynamic Kernel Learning} \label{sec: kernel learning}
Dynamic kernel learning aims at dynamically learning and searching the associations of kernels to consider the different meta-knowledge patterns in multivariate time series. The dynamic kernel learning starts from a basic kernel set $\mathcal{K}$. In Fig.~\ref{fig:DMDKL}(b), we assume $\mathcal{K}=\{k_1, ..., k_4\}$, corresponding to the four kernel functions shown in~\autoref{Base_kernel_functions}. The component automatically expands and learns potential new kernels via kernel association search (KAS) to achieve the final kernel function $K_\theta$, which is an association of basic kernels, e.g., $k_1$ and $k_2$, and expanded kernels, e.g., $k_S$. For simplicity, we denote $K_\theta$ as $K$.

Instead of using a simple and fixed kernel function to model a fixed meta-knowledge pattern among subsequences $\mathbf{M}$, which fails to capture the diversity among time series, such as combinations of growing trends and periodic patterns, we propose to learn a kernel function $K$ to contend with such diversity. It encompasses {\it kernel association}, which defines the rules of structural exploration of new kernels, and {\it kernel search}, which applies two kernel search methods to obtain the final kernel $K$.

\subsubsection{Kernel Association}
A more complex kernel function can be obtained by the addition and multiplication of existing kernel functions because positive semi-definite covariance matrices are closed under addition and multiplication~\cite{duvenaud2013structure}.
\begin{lemma}
\label{KA}
Let $k_1$ and $k_2$ be kernels and let $+$ and $\times$ denote addition and multiplication, respectively. Then $k_1 + k_2$ and $k_1\times k_2$ are kernels.
\end{lemma}

Lemma \ref{KA} allows us to expand basic kernels into high order kernels with power larger than 2. For example, LIN is a basic kernel, and LIN$^2$ is a second-order kernel. The basic kernel LIN denotes a trend pattern, while the high order kernel LIN$^2$ denotes a quadratic pattern. When a high order kernel is combined with basic kernels into a final kernel $K$, it exhibits a more complex analytic form than the basic kernels. This procedure is called kernel association~\cite{motai2014kernel}. The intuition is that kernels with different orders should be included in $K$ because basic kernels capture general patterns of time series, while higher order kernels capture higher order information. As more higher order kernels are included in $K$, $K$ approximates a real-world pattern better. This is similar to polynomial approximation~\cite{hastie2009elements}.

Specifically, we define a basic kernel set $\mathcal{K}=\{k_1,k_2,...,k_{|\mathcal{K}|}\}$, where $k_i$ is a kernel, and $|\mathcal{K}|$ is the number of kernels in $\mathcal{K}$. Then we define the operation set $\mathcal{O}=\{+,\times\}$. Our goal is to discover a suitable final kernel $K$, which is an expanded high order kernel obtained via operations in $\mathcal{O}$.

\subsubsection{Kernel Search} We procee to introduce two kernel search methods---greedy kernel search, used by existing models~\cite{grosse2012exploiting,duvenaud2013structure,lloyd2014automatic,tong2019discovering}, and KAS, a novel search method.

\begin{figure*}[htbp]
\centering
\subfigure[Greedy Kernel Search]{
\includegraphics[width=0.425\textwidth]{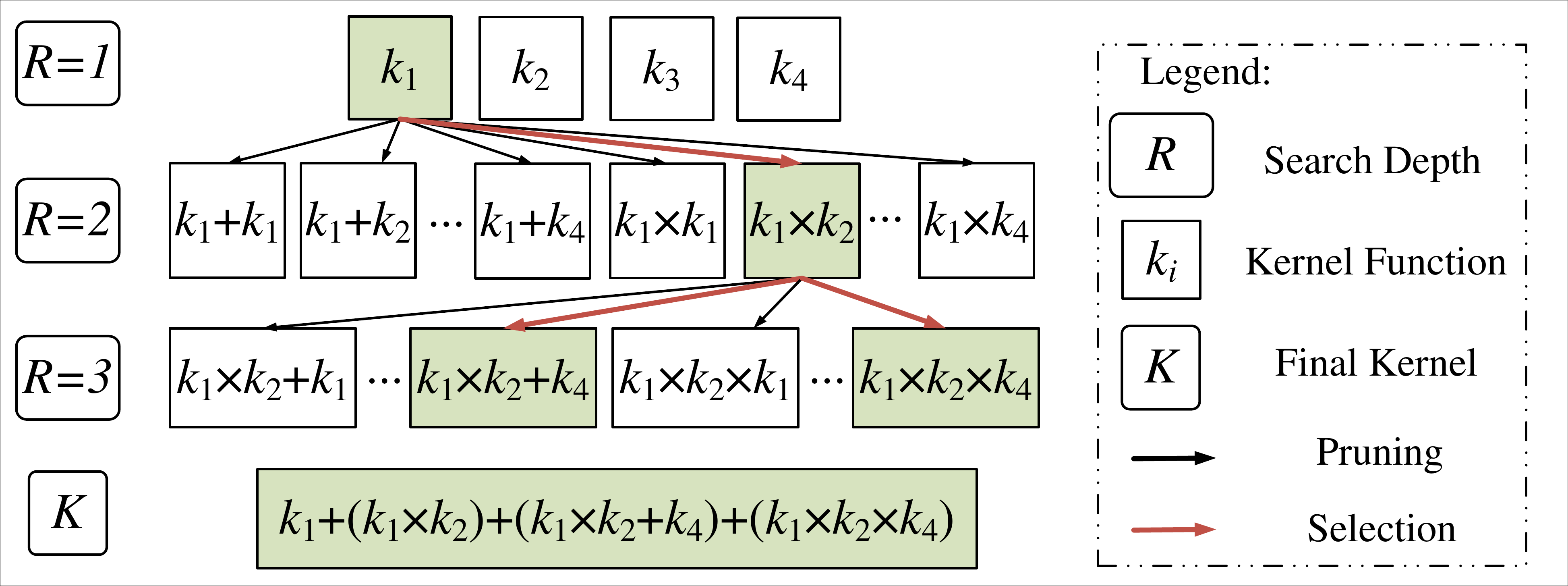}}
\subfigure[Kernel Association Search]{
\includegraphics[width=0.465\textwidth]{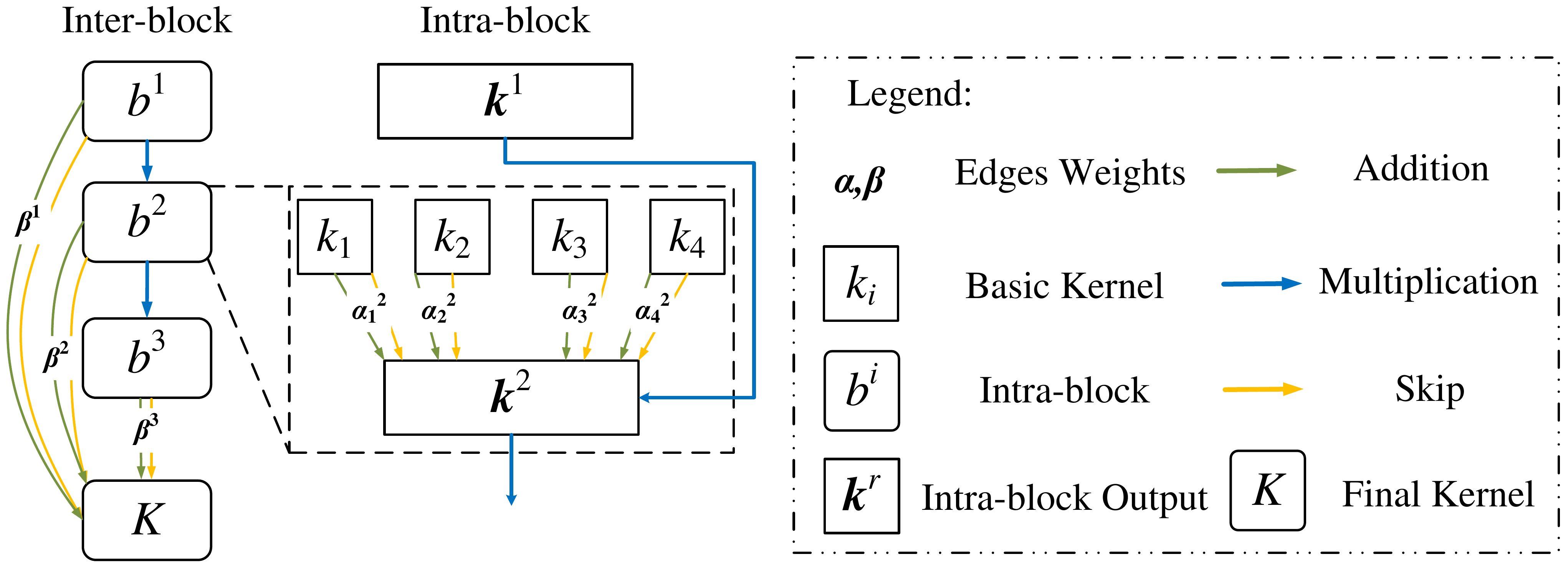}}
\caption{Kernel Search Methods.}
\label{fig:kernel_search} 
\end{figure*}

\noindent{\bf Greedy Kernel Search:} As shown in Fig.~\ref{fig:kernel_search}(a), assuming a basic kernel set $\mathcal{K}=\{k_1,k_2,k_3,k_4\}$, an operation set $\mathcal{O}=\{+,\times\}$, and a search depth $R =3$, the goal of greedy kernel search is to discover a final kernel $K$. The detailed procedure is presented in Algorithm~\ref{alg:greedy_kernel_search}.
\begin{algorithm}[htbp]
\footnotesize
\caption{Greedy Kernel Search}
\label{alg:greedy_kernel_search}
\begin{algorithmic}[1]
\renewcommand{\algorithmicrequire}{\textbf{Input:}}
\renewcommand{\algorithmicensure}{\textbf{Output:}}
\REQUIRE Kernel set $\mathcal{K}=\{k_1,k_2,\dots,k_{|\mathcal{K}|}\}$, operations $\mathcal{O}=\{+,\times\}$, depth $R$
\ENSURE Final kernel $K$

\STATE Initialize $K$ by randomly selecting $k \in \mathcal{K}$

\FOR{$r=1,\dots,R$}
    \STATE Generate candidate kernels using $\mathcal{O}$ and $\mathcal{K}$
    \STATE Select the kernel minimizing the forecasting loss and update $K$
\ENDFOR

\RETURN $K$
\end{algorithmic}
\end{algorithm}

The method starts by randomly selecting a basic kernel from kernel set $\mathcal{K}$, forming an initial kernel $K$. It then evaluates each candidate kernel formed by combining $K$ with other kernels using operations from $\mathcal{O}=\{+, \times\}$, selecting the combination that minimizes the forecasting loss $\mathcal{L}_{que}$. This procedure continues until reaching a predefined search depth $R$. In each iteration, new candidate kernels are generated by combining previously selected kernels with basic kernels, continually optimizing $K$. The final kernel $K$ is the sum of all selected kernels from each iteration.

While greedy kernel search discovers and expands kernels automatically, forecasting loss computation must be performed each time a candidate kernel is checked, causing a high time complexity of $O(|\mathcal{K}|^R)$.

\noindent{\bf Kernel Association Search:} To reduce the complexity, we propose a novel 2-level kernel search method (KAS). Inspired by neural architecture search methods~\cite{liu2018darts,wu2021autocts}, it includes intra-block and inter-block kernel associations, as shown in Fig.~\ref{fig:kernel_search}(b). Each intra-block association $b^r$ constructs an $r$-th order kernel $\bm{k}^r$, and the inter-block association sums kernels $\bm{k}^1,...,\bm{k}^R$ to obtain the final kernel $K$. The algorithm detailing the procedure is presented in Algorithm~\ref{alg:KAS}.
\begin{algorithm}[htbp]
\footnotesize
\caption{Kernel Association Search (KAS)}
\label{alg:KAS}
\begin{algorithmic}[1]
\renewcommand{\algorithmicrequire}{\textbf{Input:}}
\renewcommand{\algorithmicensure}{\textbf{Output:}}
\REQUIRE Basic kernel set $\mathcal{K}=\{k_1,k_2,...,k_{|\mathcal{K}|}\}$, edges $\mathcal{E}=\{+,\mathit{skip}\}$, intra-block number $R$
\ENSURE Final kernel $K$

\STATE \textit{Initialization}: Set $\bm{k}^0=1$

\FOR{$r=1,2,\dots,R$}
    \STATE \textit{Intra-block addition}:
    \FOR{$k_i \in \mathcal{K}$}
        \STATE Compute edge weights $\bm{\alpha}_i^r=(\alpha_{i,1}^r,\dots,\alpha_{i,|\mathcal{E}|}^r)$
        \STATE Select operation $e_j\in\mathcal{E}$ s.t. $j=\arg\max_j\alpha_{i,j}^r$
    \ENDFOR
    \STATE $\bm{k}^r=\sum_{i=1}^{|\mathcal{K}|}\sum_{j=1}^{|\mathcal{E}|}\frac{\exp(\alpha_{i,j}^r)}{\sum_{j}\exp(\alpha_{i,j}^r)}e_j(k_i)$

    \STATE \textit{Intra-block multiplication}:
    \STATE Update high-order kernel: $\bm{k}^r \leftarrow \bm{k}^{r-1}\times \bm{k}^r$
\ENDFOR

\STATE \textit{Inter-block association}:
\FOR{$r=1,2,\dots,R$}
    \STATE Compute edge weights $\bm{\beta}^r=(\beta_1^r,\dots,\beta_{|\mathcal{E}|}^r)$
    \STATE Select operation $e_j\in\mathcal{E}$ s.t. $j=\arg\max_j\beta_j^r$
\ENDFOR
\STATE Construct final kernel: $K=\sum_{r=1}^{R}\sum_{j=1}^{|\mathcal{E}|}\frac{\exp(\beta_j^r)}{\sum_{j}\exp(\beta_j^r)}e_j(\bm{k}^r)$

\STATE \textit{Kernel Weighting}:
\STATE Retrain the model and learn kernel weights $\bm{\zeta}=(\zeta_1,\dots,\zeta_S)$
\STATE Update final kernel with weights: $K=\sum_{s=1}^{S}\zeta_s k_s$

\RETURN $K$
\end{algorithmic}
\end{algorithm}

Each intra-block association $b^r$ encompasses two steps: addition and multiplication. The addition step takes basic kernels $\mathcal{K}=\{k_1,k_2,...,k_{|\mathcal{K}|}\}$ as input, producing an intermediate kernel $\bm{k}^r$ as a weighted sum based on edge operations $\mathcal{E}=\{+, \mathit{skip}\}$. The weight $\alpha_{i,j}^r$ determines whether kernel $k_i$ is included ($+$) or omitted ($\mathit{skip}$) when constructing $\bm{k}^r$. Formally, the intermediate kernel is calculated as follows.
\begin{equation}
\bm{k}^r= \sum_{i=1}^{|\mathcal{K}|}\sum_{j=1}^{|\mathcal{E}|}\frac{\mathrm{exp}(\alpha_{i,j}^r)}{\sum_{j=1}^{|\mathcal{E}|}\mathrm{exp}(\alpha_{i,j}^r)}e_{j}(k_i),
\end{equation}
where $e_j$ is the $j$-th operation in $\mathcal{E}$.

The multiplication step updates the intermediate kernel to a higher-order kernel by multiplying it with the previous kernel: $\bm{k}^r \leftarrow \bm{k}^{r-1}\times \bm{k}^r$, with $\bm{k}^0=1$.

Inter-block associations aggregate kernels from multiple intra-blocks. Each intra-block kernel $\bm{k}^r$ is integrated into the final kernel $K$ through weighted edges $\beta_j^r$, deciding inclusion or omission similarly and using operations in $\mathcal{E}$:
\begin{equation}
K=\sum_{r=1}^{R}\sum_{j=1}^{|\mathcal{E}|}\frac{\mathrm{exp}(\beta_j^r)}{\sum_{j=1}^{|\mathcal{E}|}\mathrm{exp}(\beta_j^r)}e_{j}(\bm{k}^r)
\end{equation}

Finally, kernel weights $\bm{\zeta}=(\zeta_1,\dots,\zeta_S)$ are learned by retraining the model to reflect the importance of each kernel:
\begin{equation}
K = \sum_{s=1}^{S}\zeta_s k_s,
\end{equation}
where $\zeta_s$ is the importance of kernel $k_s$.

The kernel selection is completed in one iteration without traversing the search tree as done in the greedy strategy. The 
time complexity is $O(|\mathcal{K}|R)$. KAS involves only one instance of forecasting query set loss computation, thereby reducing the time complexity.

\subsection{Multivariate Covariance Learning} \label{sec: multi}
Fig.~\ref{fig:DMDKL}(c) shows the third component in the framework, multivariate covariance learning, that aims to capture multivariate correlations among variables. Taking the Gaussian locations $\mathbf{H}$ and the learned kernel $K$ as input, the component calculates the multivariate covariance matrix $\mathbf{K}(\mathbf{H},\mathbf{H})$ among Gaussian locations $\mathbf{H}$ via the learned kernel $K$. For simplicity, we denote $\mathbf{K}(\mathbf{H}, \mathbf{H})$ by $\mathbf{K}$. We use $\mathbf{K}$ to capture the correlations between each two subsequences of different variables. 

To learn $\mathbf{K}$, we first calculate a shared-kernel covariance $\mathbf{C}(\mathbf{H},\mathbf{H})$, denoted by $\mathbf{C}$, via a multi-task Gaussian process~\cite{parra2017spectral}, a Gaussian process that handles multiple variables; and then we learn a matrix $\mathbf{D}$ that captures the cross-variable weights between each pair of variables. Next, it learns the cross-variable weight matrix $\mathbf{D}\in \mathbb{R}^{N^2}$ that captures the importance between each pair of variables. Finally, we get the multivariate covariance $\mathbf{K}$ as the Hadamard product of $\mathbf{C}$ and $\mathbf{D}$ for the correlations between any two subsequences of any two variables.

\subsubsection{Shared-kernel Covariance}
The shared-kernel covariance matrix $\mathbf{C}\in \mathbb{R}^{B^2\times N^2}$ is learned from the attention-based Gaussian locations $\mathbf{H}$ and the kernel function $K$. An element $\mathbf{C}^{m,n}\in \mathbb{R}^{B^2}$ in $\mathbf{C}$ denotes the covariance between the $m$-th and $n$-th variables in $\mathbf{H}$:
\begin{equation}
\begin{aligned}
    &\mathbf{C} =  \left[
            \begin{matrix}
             \mathbf{C}^{1,1}    & \mathbf{C}^{1,2}    & \cdots & \mathbf{C}^{1,N}         \\
             \mathbf{C}^{2,1}    & \mathbf{C}^{2,2}    & \cdots & \vdots         \\
             \vdots         & \vdots         & \ddots & \vdots    \\
             \mathbf{C}^{N,1}    & \mathbf{C}^{N,2}    & \cdots & \mathbf{C}^{N,N} \\
            \end{matrix}
        \right]
\end{aligned}
\end{equation}

Element $\mathbf{C}^{m,n}$ is also a matrix, an element $\mathbf{C}_{i,j}^{m,n}$ of which denotes the temporal covariance between the $i$-th subsequence of the $m$-th variable $\mathbf{h}_i^m$ and the $j$-th subsequence of the $n$-th variable $\mathbf{h}_j^n$. Assuming that the final kernel $K$ contains $S$ kernel functions, we have:
\begin{equation}
   \mathbf{C}_{i,j}^{m,n} =
   K(\mathbf{h}_i^m,\mathbf{h}_j^{n}) = 
   \sum_{s=1}^{S}\zeta_{s}k_{s}(\mathbf{h}_i^m,\mathbf{h}_j^{n})
\end{equation}

\subsubsection{Cross-variable Weights} The multi-task Gaussian process captures the multivariate covariance by means of a single shared kernel function $K$. As this limits the ability to capture diverse correlations between variables, we propose a learnable matrix $\mathbf{D}\in\mathbb{R}^{N^2}$ that represents the cross-variable weights, thus extending the capabilities of a single shared kernel. To ensure that covariance matrix $\mathbf{D}$ is positive semi-definite, we define $\mathbf{D}$ as follows.
\begin{equation}
\begin{aligned}
    \mathbf{D} = \mathbf{D^\prime}(\mathbf{D}^\prime)^\top + diag(\bm{\lambda}),
\end{aligned}
\end{equation}
where $\mathbf{D}^\prime\in \mathbb{R}^{N\times V}$ is a low-rank matrix, $V\ (V<N)$ represents the number of non-zero columns in $\mathbf{C}$, and $\bm{\lambda}$ is a non-negative vector. 

The multivariate covariance matrix $\mathbf{K}$ is calculated using the Hadamard product of the shared-kernel covariance and the cross-variable weights:
\begin{equation}
\begin{aligned}
    \mathbf{K} &= \sum_{s=1}^{S}\zeta_{s}
                \left[
                     \begin{matrix}
                         d_{1,1}k_{s}(\mathbf{h}^1,\mathbf{h}^1) & \cdots & d_{1,N}k_{s}(\mathbf{h}^1,\mathbf{h}^N)\\
                         d_{2,1}k_{s}(\mathbf{h}^2,\mathbf{h}^1)& \cdots & \vdots\\
                         \vdots & \ddots & \vdots\\
                         d_{N,1}k_{s}(\mathbf{h}^N,\mathbf{h}^1) & \cdots & d_{N,N}k_{s}(\mathbf{h}^N,\mathbf{h}^N)\\
                     \end{matrix}
                 \right] \\
               & = \mathbf{D} \odot \mathbf{C},
\end{aligned}
\end{equation}
where $d_{m,n}$ denotes the weight of $\mathbf{C}^{m,n}$ that constructs the cross-variable correlation between any pair of variables. Thus, the correlation between any two subsequences of two variables also considers the distinct correlation between the two variables. Matrix $\mathbf{K}$ will be fed into the multi-task Gaussian process for training and inference. 

\subsection{MetaGP Optimization}
Next, we consider parameter optimization in the MetaGP framework. As the multivariate time series forecasting is formulated as a Gaussian process, we use negative log marginal likelihood~\cite{wilson2016stochastic} rather than the general regression loss function from deep learning to optimize parameters.

The negative log marginal likelihood loss function is defined as follows.
\begin{equation}
   \mathcal{L}=-[\mathbf{y}^\top(\mathbf{K}+{\sigma}^2I)^{-1}\mathbf{y}+\log{\det|\mathbf{K}+{\sigma}^2I|}]+\mathit{const},
\end{equation}
where $\mathbf{y}$ denotes a series of observations, $\mathbf{K}$ represents a learned covariance matrix with analytic form $K$, and $\sigma^2$ is the variance of noise. Next, $\mathbf{y}^\top(\mathbf{K}+{\sigma}^2I)^{-1}\mathbf{y}$ is a positive semi-definite covariance matrix, $\mathrm{det}|\cdot|$ is the determinant of the argument matrix, and $const$ is a constant term. 

The framework learns the analytic form of kernel function $K$ achieved via KAS and parameters $\bm{\Gamma}=\{\mathbf{w},K,\bm{\theta},\mathbf{D}\}$, where $\mathbf{w}$ denotes the learnable weights in the patch-attention, $\bm{\theta}$ represents the parameters of kernel $K$, which includes $\bm{\alpha}$, $\bm{\beta}$, $\bm{\zeta}$, and other parameters related to kernels, and $\mathbf{D}$ is the cross-variable weights. Specifically, the cross-variable weights $\mathbf{D}$ is a positive semi-definite matrix. As kernel computation is closed under the Hadamard product operation, the optimization of $\mathbf{D}$ and $\bm{\theta}$ represents the parameter optimization of kernel $K$. Therefore, $\mathbf{D}$ is regarded as a covariance matrix, and the optimization procedures of $\mathbf{D}$ and $\bm{\theta}$ are combined. We let $\bm{\Theta}=\{ \bm{\theta},\mathbf{D}\}$. The framework then learns parameters $\bm{\Theta}$ and $\mathbf{w}$ jointly via back-propagation using loss function $\mathcal{L}$ and gradient descent.
The partial derivative $\frac{\partial{\mathcal{L}}}{\partial{\bm{\Theta}}}$ is defined as follows.
\begin{equation}
   \frac{\partial{\mathcal{L}}}{\partial{\bm{\Theta}}} = \frac{1}{2}\mathrm{tr}\left(\left[\mathbf{K}^{-1}\mathbf{y}\mathbf{y}^{\top}\mathbf{K}^{-1} - \mathbf{K}^{-1}\right]\frac{\partial{\mathbf{K}}}{\partial{\bm{\Theta}}}\right)
\end{equation}

The other partial derivative $\frac{\partial{\mathcal{L}}}{\partial{\mathbf{w}}}$ is defined as follows.
\begin{small}
\begin{equation}
\begin{aligned}
   \frac{\partial{\mathcal{L}}}{\partial{\mathbf{w}}} =& \frac{1}{2}\sum_{m,n}\sum_{i,j}(\mathbf{K}(\mathbf{h}_i^m, \mathbf{h}_j^{n})^{-1}\mathbf{y}\mathbf{y}^{\top}\mathbf{K}(\mathbf{h}_i^m, \mathbf{h}_j^{n})^{-1} - \mathbf{K}(\mathbf{h}_i^m, \mathbf{h}_j^{n})^{-1}) \\
   &\left\{\left(\frac{\partial{\mathbf{K}(\mathbf{h}_i^m, \mathbf{h}_j^{n})}}{\partial{\mathbf{h}^m_i}}\right)^{\top}\frac{\partial{\mathbf{h}^m_i}}{\partial{\mathbf{w}}} + \left(\frac{\partial{\mathbf{K}(\mathbf{h}_i^m, \mathbf{h}_j^{n})}}{\partial{\mathbf{h}^n_j}}\right)^{\top}\frac{\partial{\mathbf{h}^n_j}}{\partial{\mathbf{w}}}\right\}
\end{aligned}
\end{equation}
\end{small}

Then we update the parameters using gradient descent as follows.
\begin{equation}
\begin{aligned}
   \bm{\Theta} = \bm{\Theta} - \eta\frac{\partial{\mathcal{L}}}{\partial{\bm{\Theta}}},\quad
   \mathbf{w} = \mathbf{w} - \eta\frac{\partial{\mathcal{L}}}{\partial{\mathbf{w}}},
\end{aligned}
\end{equation}
where $\eta$ is a learning rate that is used to control the learning speed.

Given a time series of length $T$ with $N$ variables, the time and space complexity of attention-based Gaussian location learning are $O(NT)$. Dynamic kernel learning has time complexity $O(|\mathcal{K}|RNT)$ and space complexity $O(NT)$. Leveraging KISS-GP~\cite{wilson2015kernel}, multivariate covariance learning reduces both complexities to $O(NT)$. Thus, the overall time and space complexity of MetaGP are $O(|\mathcal{K}|RNT)$ and $O(NT)$, respectively.
\section{Experimental Study} 
\label{experiments}

\subsection{Experimental Setup}
\subsubsection{Datasets and Evaluation Metrics} We perform experimental studies on seven benchmark datasets.

\noindent\textbf{Gravity-16\footnote[1]{\href{https://drive.google.com/drive/folders/1Tm3DNrugcSbWXSNyeGL3jQKR8y3iXx0m}{Gravity-16 Dataset}}:} The Gravity-16 dataset~\cite{jiang2022sequential} consists of a sequence of image frames capturing the trajectory of a ball under varying initial positions and velocities across 16 different directions of gravity. These continuous images form time-series data. \\
\textbf{Mixed-physics\footnote[2]{\href{https://drive.google.com/drive/folders/1Tm3DNrugcSbWXSNyeGL3jQKR8y3iXx0m/}{Mixed-physics Dataset}}:} The Mixed-physics dataset comprises three sub-datasets: Bouncing Ball~\cite{fraccaro2017disentangled}, Pendulum, and Mass-spring~\cite{botev2021priors}. The Bouncing Ball dataset describes sequences of image frames capturing the trajectory of a ball under 4 different directions of gravity. Pendulum and Mass-spring are also image sequence datasets, now capturing the dynamics of pendulums and mass springs under 4 different friction coefficients.  \\
\textbf{MC\footnote[3]{\href{https://github.com/KasperSkytte/MC-prediction/tree/main}{MC Dataset}}:} The MC dataset, contributed by the Department of Chemistry and Bioscience at Aalborg University, describes a time series of microbial community scales in activated sludge sampled from different regions.
\begin{table}[htbp]
\setlength{\tabcolsep}{1.2mm}{}{
\caption{Dataset Statistics.}
\begin{tabular}{ccccccccc}
\hline
Dataset       & Tasks & Samples & $N$  & Length & Ratio & Shot & $L$ & $F$ \\ \hline
Gravity-16    & 16    & 3000    & 1024 & 100    & 5:1:2 & 15   & 8   & 12  \\
Bouncing Ball & 4     & 3000    & 1024 & 100    & 3:0:1 & 15   & 8   & 12  \\
Pendulum      & 4     & 3000    & 1024 & 100    & 3:0:1 & 15   & 8   & 12  \\
Mass-spring   & 4     & 3000    & 1024 & 100    & 3:0:1 & 15   & 8   & 12  \\
MC            & 24    & 177     & 951  & 196    & 5:1:2 & 15   & 8   & 12  \\ \hline
\end{tabular}
\label{datasets}
}
\end{table}

\autoref{datasets} provides statistics of the benchmark datasets, where "Tasks" refers to the total number of tasks utilized for meta-training, meta-validation, and meta-testing, with each task consisting of a dataset formed by a group of time series with similar distributions, ``Samples'' refers to the average number of time series for all tasks, $N$ denotes the average number of variables, ``Length'' represents the average number of observations, ``Ratio'' indicates the ratio of meta training-validation-testing tasks for the entire dataset, ``shot'' indicates the number of samples used for few-shot learning, and $L$ and $F$ denote the lengths of the historical and forecasting horizons, respectively.

On the Gravity-16 and Mixed-physics datasets, MetaGP aligns the dataset statistics strictly with those reported by Jiang et al.~\cite{jiang2022sequential}. Specifically, in Gravity-16, the "Ratio" 5:1:2 indicates that MetaGP randomly selects 10 tasks for meta-training, 2 tasks for meta-validation, and 4 tasks for meta-testing. In meta-learning, for each task, the support set consists of ``shot'' randomly selected samples, and the remaining samples are partitioned into the query set. For the MC, we align the ``Ratio'', ``Shot'', $L$, and $F$ with those of Gravity-16. \\
\noindent {\bf Metrics:} We evaluate prediction accuracy using four metrics~\cite{jiang2022sequential}: distance (Dist), valid prediction time distance (VPT-Dist), mean absolute error (MAE), and mean squared error (MSE). In image sequences, MAE and MSE are challenging to use for measuring the error between prediction and ground truth, as the pixel-level moving objects are very small. Therefore, Jiang et al.~\cite{jiang2022sequential} propose the metrics Dist and VPT-Dist. Specifically, Dist quantifies the distance between the predicted object trajectory and the ground truth trajectory, while VPT-Dist assesses the duration during which the predicted object's trajectory aligns closely with the ground truth trajectory. Additionally, we adopt the commonly used metrics MAE and MSE~\cite{nie2022time} for the biological dataset MC.

\subsubsection{Baselines} We consider 6 categories of baselines. {\it System state as latent variable}~\cite{chung2015recurrent,krishnan2017structured} do not consider parameters as priors that provide additional information to the model; these include VRNN and DKF. (1) VRNN~\cite{chung2015recurrent} is a traditional sequential latent variable model that incorporates latent random variables into the hidden state of a recurrent neural network. (2) DKF~\cite{krishnan2017structured} is a Gaussian-based state space model that introduces a unified algorithm to learn a broad class of linear and non-linear state space models. {\it System parameter as latent variable}~\cite{karl2022deep,fraccaro2017disentangled,jiang2022sequential,botev2021priors} consider parameters as priors; these include DVBF, KVAE, GRU-res, NODE, and RGN-res. (3) DVBF~\cite{karl2022deep} is a method for unsupervised learning and identification of latent Markovian state space models, which treats system parameters as latent variables. (4) KVAE~\cite{fraccaro2017disentangled}, a Kalman variational auto-encoder, incorporates system parameters into the process of sequential latent variable inference. (5) GRU-res~\cite{jiang2022sequential} is a sequential LVM that incorporates external parameters as priors and employs residual gated recurrent units as the latent dynamic function. (6) NODE~\cite{jiang2022sequential} is similar to GRU-res, but utilizes the neural ordinary differential equation as the latent dynamic function. (7) RGN-res~\cite{botev2021priors} is similar to GRU-res, but utilizes the residual recurrent generative networks as the latent dynamic function. {\it Meta-learning LVMs}~\cite{jiang2022sequential} integrate meta-learning theories into sequential latent variable models; these include meta-GRU-res, meta-NODE, and meta-RGN-res. (8) Meta-GRU-res is the meta-learning version of GRU-res, which shows advantages at few-shot learning on image sequences. (9) Meta-NODE is the meta-learning version of NODE. (10) Meta-RGN-res is the meta-learning version of RGN-res. {\it Autoregressive model}~\cite{dona2020pde} is included to evaluate the performance of autoregressive models in few-shot learning. (11) Don\`{a} et al. present an autoregressive model designed to tackle forecasting according to diverse dynamics. \textit{Gaussian Process}~\cite{lalchand2022generalised,williams2006gaussian} involves methods based on Gaussian processes. (12) GP-LVM~\cite{lalchand2022generalised} is a Gaussian Process Latent Variable Model. (13) GP~\cite{williams2006gaussian} refers to the traditional Gaussian Process. \textit{Meta-learning}~\cite{wang2022meta,oreshkin2021meta} involves methods based on meta-learning. (14) DyAd~\cite{wang2022meta} is a meta-learning-based dynamic adaptation network. (15) Meta-N-Beats~\cite{oreshkin2021meta} is the meta-learning version of N-BEATS.

\subsubsection{Implementation Details}
All experiments are conducted on a server running Linux 18.04 with an Intel Xeon W-2155 CPU @ 3.30GHz and two RTX GPUs with 24GB memory. 

We use the Adam~\cite{kingma2015ba} optimizer with a learning rate $\eta$ in the range $[1e^{-5},1e^{-2}]$. The number of training epochs is chosen from $\{50,100,200,500\}$, taking into account the capacities of the datasets. The basic kernel set includes four simple kernel functions: SE, RQ, LIN, and PER. The candidate patch-size is chosen from all the common divisors of the length of the time window. The order $R$ is chosen among $\{1,2,3,4,5\}$. The moving step $\Delta$ is choosen among $\{1,2,4,8\}$. We use grid search~\cite{pontes2016design} to select optimal hyperparameters from among the candidates described above. Furthermore, we tune the hyperparameters carefully based on the recommendations provided for the baselines. 
\begin{table*}[htpb]
\footnotesize
\caption{Comparison of Prediction Accuracy on Gravity-16 and MC.}
\centering
\label{main_exp}
\setlength\tabcolsep{10pt}
\begin{tabular}{cc|cc|cc}
\hline
\multicolumn{2}{c|}{Dataset}                                                                                                         & \multicolumn{2}{c|}{Gravity-16}           & \multicolumn{2}{c}{MC}                          \\ \hline
Model Category                                                                                         & Model                       & Dist$\downarrow$    & VPT-Dist$\uparrow$  & MAE$\downarrow$        & MSE$\downarrow$        \\ \hline
Ours                                                                                                   & MetaGP                      & \textbf{2.06 (0.85)} & \textbf{0.99 (0.05)} & \textbf{1.21 (0.14)e-1} & \textbf{0.32 (0.03)e-1} \\ \hline
\multirow{3}{*}{Meta-learning LVMs}                                                                    & meta-GRU-res                & 2.88 (1.45)          & 0.97 (0.07)          & 1.38 (0.18)e-1          & 0.46 (0.05)e-1          \\
                                                                                                       & meta-Node-res               & 6.10 (2.63)          & 0.80 (0.12)          & 2.24 (0.16)e-1          & 0.78 (0.12)e-1          \\
                                                                                                       & meta-RGN-res                & 6.97 (3.08)          & 0.76 (0.13)          & 2.41 (0.16)e-1          & 0.84 (0.13)e-1          \\ \hline
\multirow{9}{*}{\begin{tabular}[c]{@{}c@{}}.\\ \\ System parameter \\ as latent variable\end{tabular}} & GRU-res                     & 10.4 (3.30)          & 0.61 (0.90)          & 3.35 (0.22)e-1          & 1.94 (0.22)e-1          \\
                                                                                                       & GRU-res finetune            & 9.35 (3.33)          & 0.66 (0.12)          & 2.64 (0.20)e-1          & 1.07 (0.18)e-1          \\
                                                                                                       & NODE                        & 10.9 (3.32)          & 0.59 (0.08)          & 3.54 (0.25)e-1          & 2.72 (0.37)e-1          \\
                                                                                                       & NODE finetune               & 10.4 (3.23)          & 0.61 (0.09)          & 2.76 (0.25)e-1          & 1.15 (0.20)e-1          \\
                                                                                                       & RGN-res                     & 11.2 (3.39)          & 0.58 (0.09)          & 3.63 (0.21)e-1          & 1.99 (0.29)e-1          \\
                                                                                                       & RGN-res finetune            & 10.0 (3.36)          & 0.62 (0.11)          & 2.74 (0.25)e-1          & 1.01 (0.19)e-1          \\
                                                                                                       & DVBF                        & 45.3 (0.00)          & 0.00 (0.00)          & 4.57 (0.49)e-1          & 3.21 (0.51)e-1          \\
                                                                                                       & DVBF finetune               & 45.3 (0.00)          & 0.00 (0.00)          & 4.48 (0.46)e-1          & 3.35 (0.55)e-1          \\
                                                                                                       & KVAE                        & 4.81 (3.61)          & 0.57 (0.29)          & 3.22 (0.15)e-1          & 2.31 (0.43)e-1          \\ \hline
\multirow{5}{*}{\begin{tabular}[c]{@{}c@{}}System state\\ as latent variable\end{tabular}}             & meta-DKF                    & 7.35 (3.26)          & 0.70 (0.25)          & 3.93 (0.53)e-1          & 2.51 (0.47)e-1          \\
                                                                                                       & DKF                         & 7.39 (3.21)          & 0.69 (0.25)          & 3.98 (0.53)e-1          & 2.53 (0.45)e-1          \\
                                                                                                       & DKF finetune                & 7.51 (3.26)          & 0.69 (0.25)          & 3.95 (0.53)e-1          & 2.53 (0.45)e-1          \\
                                                                                                       & VRNN                        & 23.1 (21.6)          & 0.51 (0.19)          & 4.24 (0.58)e-1          & 3.46 (0.51)e-1          \\
                                                                                                       & VRNN finetune               & 8.31 (11.6)          & 0.75 (0.19)          & 3.37 (0.62)e-1          & 2.71 (0.45)e-1          \\ \hline
Autoregressive                                                                                         & Don\`a et al. & 13.7 (3.05)          & 0.06 (0.15)          & 4.06 (0.19)e-1          & 3.34 (0.50)e-1          \\ \hline
\multirow{2}{*}{Gaussian Process}                                                                      & GP-LVM                      & 4.63 (2.61)          & 0.74 (0.15)          & 2.32 (0.15)e-1          & 1.07 (0.23)e-1          \\
                                                                                                       & GP                          & 8.81 (3.79)          & 0.58 (0.17)          & 4.70 (0.32)e-1          & 2.24 (0.49)e-1          \\ \hline
\multirow{2}{*}{Meta-Learning}                                                                         & DyAd                        & 3.05 (0.98)          & 0.85 (0.12)          & 2.45 (0.19)e-1          & 0.63 (0.13)e-1          \\
                                                                                                       & Meta-N-BEATS                & 4.35 (1.23)          & 0.78 (0.14)          & 2.52 (0.18)e-1          & 0.74 (0.13)e-1          \\ \hline
\end{tabular}
\end{table*}

\subsection{Comparison of Prediction Accuracy}
We proceed to study prediction accuracy on Gravity-16 and MC. Standard deviations are provided in parentheses. The best results are highlighted in boldface.

\autoref{main_exp} shows a comparison of the average prediction accuracy and standard deviation across five repeated experiments for MetaGP and the baselines. The results show that MetaGP achieves state-of-the-art prediction accuracy on Gravity-16 and MC.

In few-shot learning, autoregressive methods exhibit lower prediction accuracy compared to meta-learning methods. This is attributed to the autoregressive nature of the methods, which necessitates a large number of samples to establish correspondences between historical and forecasting horizons within latent spaces. Consequently, autoregressive methods struggle to generalize model parameters to new tasks using only a limited number of samples.

Both methods, whether utilizing system parameters as latent states or system states as latent states, perform worse than the meta-learning methods on Gravity-16. However, on the MC, the approach employing system parameters as latent states outperforms the approach using system states as latent states. This is attributed to the presence of long-term dependencies in the MC time series when compared to the Gravity-16 image sequences. Using parameters as priors provides global information, which helps alleviate the challenges faced by sequential LVMs in handling long-term dependencies. In addition, the fine-tuned versions of the baselines show slight improvements in prediction accuracy over their non-fine-tuned counterparts. This suggests that although simple fine-tuning methods can mitigate the issue of insufficient samples, the effect of fine-tuning is minimal.

Furthermore, Gaussian process-based methods achieve limited prediction accuracy due to their reliance on a single prior. Next, as in the case of LVMs, meta-learning-based methods achieve limited prediction accuracy due to their limitations in capturing long-term dependencies.

The meta-learning LVMs exhibit prediction accuracies that are second only to that of MetaGP on the two datasets, indicating that the meta-learning method benefits sequential LVMs in few-shot learning. Notably, the decline in prediction accuracy of meta-learning LVMs is more pronounced on MC than on Gravity-16. This suggests that the Gaussian processes help MetaGP capture long-term dependencies.
\begin{table*}[htbp]
\setlength\tabcolsep{5.4pt}
\footnotesize
\centering
\caption{Ablation Study of MetaGP.}
\label{ablation_study}
\setlength\tabcolsep{10pt}
\begin{tabular}{cc|cc|cc}
\hline
\multicolumn{2}{c|}{Dataset}                               & \multicolumn{2}{c|}{Gravity-16}           & \multicolumn{2}{c}{MC}                          \\ \hline
\multicolumn{1}{c|}{Model}                           & Horizon $F$ & Dist$\downarrow$    & VPT-Dist$\uparrow$  & MAE$\downarrow$        & MSE$\downarrow$        \\ \hline
\multicolumn{1}{c|}{\multirow{3}{*}{MetaGP}}         & 12  & \textbf{2.06 (0.85)} & \textbf{0.99 (0.05)} & \textbf{1.21 (0.14)e-1} & \textbf{0.32 (0.03)e-1} \\
\multicolumn{1}{c|}{}                                & 24  & \textbf{2.34 (0.94)} & \textbf{0.98 (0.06)} & \textbf{1.26 (0.19)e-1} & \textbf{0.34 (0.04)e-1} \\
\multicolumn{1}{c|}{}                                & 48  & \textbf{3.02 (1.59)} & \textbf{0.95 (0.08)} & \textbf{1.39 (0.21)e-1} & \textbf{0.37 (0.04)e-1} \\ \hline
\multicolumn{1}{c|}{\multirow{3}{*}{MetaGP w/o Att}} & 12  & 2.14 (1.24)          & 0.98 (0.07)          & 1.27 (0.15)e-1          & 0.39 (0.04)e-1          \\
\multicolumn{1}{c|}{}                                & 24  & 2.85 (1.52)          & 0.95 (0.09)          & 1.36 (0.24)e-1          & 0.41 (0.05)e-1          \\
\multicolumn{1}{c|}{}                                & 48  & 3.39 (1.82)          & 0.94 (0.09)          & 1.53 (0.25)e-1          & 0.45 (0.05)e-1          \\ \hline
\multicolumn{1}{c|}{\multirow{3}{*}{MetaGP w/o GP}}  & 12  & 2.54 (1.43)          & 0.96 (0.08)          & 1.33 (0.22)e-1          & 0.41 (0.05)e-1          \\
\multicolumn{1}{c|}{}                                & 24  & 3.98 (2.32)          & 0.85 (0.10)          & 2.03 (0.26)e-1          & 1.06 (0.18)e-1          \\
\multicolumn{1}{c|}{}                                & 48  & 5.72 (3.39)          & 0.79 (0.13)          & 2.43 (0.30)e-1          & 1.91 (0.22)e-1          \\ \hline
\multicolumn{1}{c|}{\multirow{3}{*}{MetaGP w/o CVW}} & 12  & 3.31 (0.93)          & 0.92 (0.02)          & 1.81 (0.25)e-1          & 0.85 (0.15)e-1          \\ 
\multicolumn{1}{c|}{}                                & 24  & 3.44 (1.21)          & 0.89 (0.03)          & 1.88 (0.25)e-1          & 1.08 (0.17)e-1          \\
\multicolumn{1}{c|}{}                                & 48  & 3.70 (1.24)          & 0.88 (0.03)          & 2.05 (0.27)e-1          & 1.10 (0.17)e-1          \\ \hline
\multicolumn{1}{c|}{\multirow{3}{*}{MetaGP w/o KAS}} & 12  & 3.05 (1.56)          & 0.97 (0.03)          & 2.04 (0.28)e-1          & 1.05 (0.16)e-1          \\
\multicolumn{1}{c|}{}                                & 24  & 3.15 (1.68)          & 0.95 (0.05)          & 2.09 (0.28)e-1          & 1.07 (0.11)e-1          \\
\multicolumn{1}{c|}{}                                & 48  & 3.42 (1.83)          & 0.92 (0.06)          & 2.26 (0.30)e-1          & 1.10 (0.13)e-1          \\ \hline
\multicolumn{1}{c|}{\multirow{3}{*}{MetaGP w/o NAS}} & 12  & 2.25 (1.03)          & 0.98 (0.06)          & 1.22 (0.15)e-1          & 0.37 (0.04)e-1          \\
\multicolumn{1}{c|}{}                                & 24  & 2.41 (1.08)          & 0.97 (0.07)          & 1.29 (0.21)e-1          & 0.40 (0.05)e-1          \\
\multicolumn{1}{c|}{}                                & 48  & 3.48 (1.65)          & 0.94 (0.07)          & 1.46 (0.23)e-1          & 0.43 (0.06)e-1          \\ \hline
\end{tabular}
\end{table*}

\subsection{Ablation Study}
We conduct an ablation study on Gravity-16 and MC to assess the effectiveness of the components in MetaGP. The results are shown in~\autoref{ablation_study}, where MetaGP w/o Att refers to MetaGP with the attention mechanism replaced by the GRU-res structure, MetaGP w/o GP denotes MetaGP with the Gaussian process replaced by a traditional sequential LVM structure, MetaGP w/o CVW indicates MetaGP without cross-variable weights and using instead a unified kernel function for all variables, MetaGP w/o KAS represents MetaGP without the meta-learning structure and using instead an SE kernel, and MetaGP w/o NAS is MetaGP without the neural architecture search-based meta-learning strategy, employing instead a greedy strategy. We fix the historical horizon at 8 but vary the forecasting horizon among 12, 24, and 48 to assess the performance of MetaGP at capturing long-term dependencies.

MetaGP w/o Att shows a slight decrease in prediction accuracy compared to MetaGP on both datasets, indicating that the attention mechanism is better at capturing multi-scale information in time series than in recurrent structures. However, the slight improvement also suggests that addressing the key challenges of few-shot time series forecasting cannot be achieved solely by replacing the encoder of the model.

MetaGP w/o GP suffers a notable decrease in prediction accuracy as forecasting horizon $F$ increases. This indicates that Gaussian processes are crucial for capturing long-term dependencies. Gaussian processes incur less accumulation of losses compared to traditional sequential LVMs. Moreover, compared to w/o Att, w/o GP suffers a much larger decline in prediction accuracy as the forecasting horizon increases, highlighting that the kernel functions of Gaussian processes play a critical role in capturing long-term dependencies. In contrast, the attention mechanism contributes primarily to strong feature representation rather than to modeling long-term temporal structures in few-shot learning.

MetaGP w/o CVW shows lower prediction accuracy than MetaGP, indicating that cross-variable weights play a crucial role in capturing correlations among multiple variables, which is beneficial for few-shot learning. Additionally, we observe that MetaGP w/o CVW has a smaller prediction standard deviation than other baselines, suggesting that the model, using a fixed kernel function, experiences underfitting. Its predictions are smoother but also less accurate.

MetaGP w/o KAS shows a more pronounced decrease in accuracy on MC compared to Gravity-16. Unlike the gravity system's image sequences, MC features more complex temporal dynamics that consist of different combinations of temporal patterns. Removing KAS limits MetaGP to using a single kernel function to capture such intricate temporal dynamics, thereby reducing MetaGP's ability to effectively model such complex information.

MetaGP w/o NAS shows prediction accuracy similar to MetaGP because both methods tend to converge to similar kernel functions during kernel search. However, the greedy strategy carries the risk of getting stuck in local optima. For instance, if a task lacks a trend pattern but the basic kernel function set includes a trend kernel, the model might retain this kernel in the initial training rounds due to reduced loss. Thus, the greedy strategy may preserve unnecessary components.

In summary, all components contribute to improving the prediction accuracy of MetaGP, demonstrating the efficacy of the framework and its components.

\subsection{Few-shot Learning and Scalability}
We assess the scalability of MetaGP by examining the impact of varying the number of few-shot learning training samples on its prediction accuracy, as well as its prediction accuracy on systems with different dynamic complexities.
\begin{table*}[htbp]
\footnotesize
\centering
\caption{Few-shot Learning of MetaGP.}
\label{few-shot}
\setlength\tabcolsep{10pt}
\begin{tabular}{cc|cc|cc}
\hline
\multicolumn{2}{c|}{Dataset}                              & \multicolumn{2}{c|}{Gravity-16}           & \multicolumn{2}{c}{MC}                          \\ \hline
\multicolumn{1}{c|}{Model}                         & Shot & Dist$\downarrow$    & VPT-Dist$\uparrow$  & MAE$\downarrow$        & MSE$\downarrow$        \\ \hline
\multicolumn{1}{c|}{\multirow{4}{*}{MetaGP}}       & 1    & \textbf{6.38 (2.60)} & \textbf{0.74 (0.07)} & \textbf{2.93 (0.21)e-1} & \textbf{1.39 (0.22)e-1} \\
\multicolumn{1}{c|}{}                              & 5    & \textbf{2.64 (1.04)} & \textbf{0.97 (0.06)} & \textbf{1.55 (0.19)e-1} & \textbf{0.30 (0.05)e-1} \\
\multicolumn{1}{c|}{}                              & 10   & \textbf{2.24 (0.89)} & \textbf{0.98 (0.05)} & \textbf{1.28 (0.15)e-1} & \textbf{0.36 (0.04)e-1} \\
\multicolumn{1}{c|}{}                              & 15   & \textbf{2.06 (0.85)} & \textbf{0.99 (0.05)} & \textbf{1.21 (0.14)e-1} & \textbf{0.32 (0.03)e-1} \\ \hline
\multicolumn{1}{c|}{\multirow{4}{*}{meta-GRU-res}} & 1    & 10.6 (3.40)          & 0.60 (0.10)          & 3.31 (0.51)e-1          & 2.49 (0.37)e-1          \\
\multicolumn{1}{c|}{}                              & 5    & 3.49 (1.89)          & 0.94 (0.10)          & 1.75 (0.24)e-1          & 1.72 (0.20)e-1          \\
\multicolumn{1}{c|}{}                              & 10   & 3.08 (1.58)          & 0.96 (0.08)          & 1.52 (0.20)e-1          & 1.64 (0.16)e-1          \\
\multicolumn{1}{c|}{}                              & 15   & 2.88 (1.45)          & 0.97 (0.07)          & 1.38 (0.18)e-1          & 0.46 (0.05)e-1          \\ \hline
\end{tabular}
\end{table*}

\subsubsection{Study of Few-shot Learning} 
We select and compare the prediction accuracy of MetaGP and the second-best meta-learning method, meta-GRU-res, on Gravity-16 and MC in few-shot learning. We set the number of few-shot learning samples to 1, 5, 10, and 15, and observe the changes in prediction accuracy.

As shown in~\autoref{few-shot}, MetaGP outperforms the baseline in prediction accuracy across all shots. Additionally, as expected, the prediction accuracy of MetaGP and the baseline increase with the number of training samples. When the number of training samples reaches 5, the rate of increase in prediction accuracy for both methods slows down. This indicates that MetaGP's performance is not significantly affected even with a small number of training samples. MetaGP's scalability and flexibility at predicting time series with any given few-shot size.
\begin{table*}[htbp]
\footnotesize
\centering
\caption{Comparison of Prediction Accuracy on the Mixed-physics Dataset.}
\label{dynamics}
\setlength\tabcolsep{10pt}
\begin{tabular}{c|cc|cc|cc}
\hline
Dataset                     & \multicolumn{2}{c|}{Bouncing Ball}        & \multicolumn{2}{c|}{Pendulum}             & \multicolumn{2}{c}{Mass-spring}           \\ \hline
Model                       & Dist$\downarrow$    & VPT-Dist$\uparrow$  & Dist$\downarrow$    & VPT-Dist$\uparrow$  & Dist$\downarrow$    & VPT-Dist$\uparrow$  \\ \hline
MetaGP                      & \textbf{2.45 (1.22)} & \textbf{0.97 (0.05)} & \textbf{1.51 (1.64)} & \textbf{0.97 (0.08)} & \textbf{0.15 (0.07)} & \textbf{1.00 (0.00)} \\
meta-GRU-res                & 4.39 (2.50)          & 0.89 (0.13)          & 1.74 (1.99)          & 0.96 (0.10)          & 0.18 (0.09)          & 1.00 (0.00)          \\
GRU-res                     & 4.61 (2.68)          & 0.89 (0.14)          & 3.57 (3.86)          & 0.87 (0.20)          & 0.25 (0.15)          & 1.00 (0.00)          \\
GRU-res finetune            & 4.37 (2.49)          & 0.90 (0.13)          & 3.55 (4.05)          & 0.87 (0.20)          & 0.16 (0.08)          & 1.00 (0.00)          \\
meta-DKF                    & 7.28 (3.55)          & 0.72 (0.26)          & 3.49 (2.96)          & 0.93 (0.18)          & 0.54 (0.29)          & 1.00 (0.00)          \\
DKF                         & 9.78 (3.45)          & 0.38 (0.26)          & 6.55 (4.21)          & 0.66 (0.35)          & 0.86 (0.48)          & 1.00 (0.00)          \\
DKF finetune                & 9.87 (3.37)          & 0.37 (0.26)          & 6.03 (4.12)          & 0.68 (0.35)          & 0.83 (0.47)          & 1.00 (0.00)          \\
KAVE                        & 10.0 (3.01)          & 0.42 (0.34)          & 4.12 (4.26)          & 0.70 (0.34)          & 0.52 (0.51)          & 1.00 (0.00)          \\
Don\`a et al. & 13.8 (2.97)          & 0.06 (0.17)          & 15.8 (2.31)          & 0.03 (0.08)          & 5.60 (2.79)          & 0.61 (0.45)          \\
GP-LVM                      & 4.49 (2.73)          & 0.90 (0.15)          & 1.89 (1.87)          & 0.93 (0.12)          & 0.24 (0.10)          & 1.00 (0.00)          \\
DyAd                        & 4.86 (2.54)          & 0.82 (0.20)          & 2.03 (2.14)          & 0.85 (0.15)          & 0.26 (0.13)          & 1.00 (0.00)          \\ \hline
\end{tabular}
\end{table*}

{\subsubsection{Study on Dynamics of Different Complexity}
We proceed to study prediction accuracy on Mixed-physics to assess the ability of MetaGP and the baselines at capturing varying meta-knowledge patterns in systems with different dynamic complexities. Jiang et al.~\cite{jiang2022sequential} find that Bouncing Ball in the Mixed-physics dataset exhibits more complex dynamics than do Pendulum and Mass-spring. This is because Bouncing Ball's underlying gravity system is more complex than those underlying Pendulum and Mass-spring.

We align our experimental setup with that of Jiang et al.~\cite{jiang2022sequential} and select one representative baseline from each of the 6 categories: meta-GRU-res, KAVE, DKF, Donà et al, GL-LVM, and DyAd. In addition, we include derived fine-tuned and meta-learning versions of these baselines, if applicable: GRU-res, GRU-res finetune, DKF finetune, and meta-DKF. The findings, in~\autoref{dynamics}, show that MetaGP achieves state-of-the-art prediction accuracy on both the complex Bouncing Ball and the simpler Pendulum and Mass-spring.

All methods exhibit higher prediction accuracy on data from simpler systems than from complex systems, indicating that as system complexity increases, the difficulty of capturing varying patterns also increases. Additionally, the fine-tuned and meta-learning versions of the baselines outperform the original versions. To mitigate the impact of the few-shot learning, we compare MetaGP with fine-tuned and meta-learning versions of the baselines. We observe that the performance improvement of MetaGP on complex systems is much better than that on simpler systems, suggesting that the meta-learning component can transform varying patterns into meta-knowledge and capture complex dynamics effectively. This indicates that MetaGP can scale to data from systems with complex dynamics.
\begin{figure*}[htbp]
\centering
\subfigure[MAE Vary with $\Delta$]{
\includegraphics[width=0.231\textwidth]{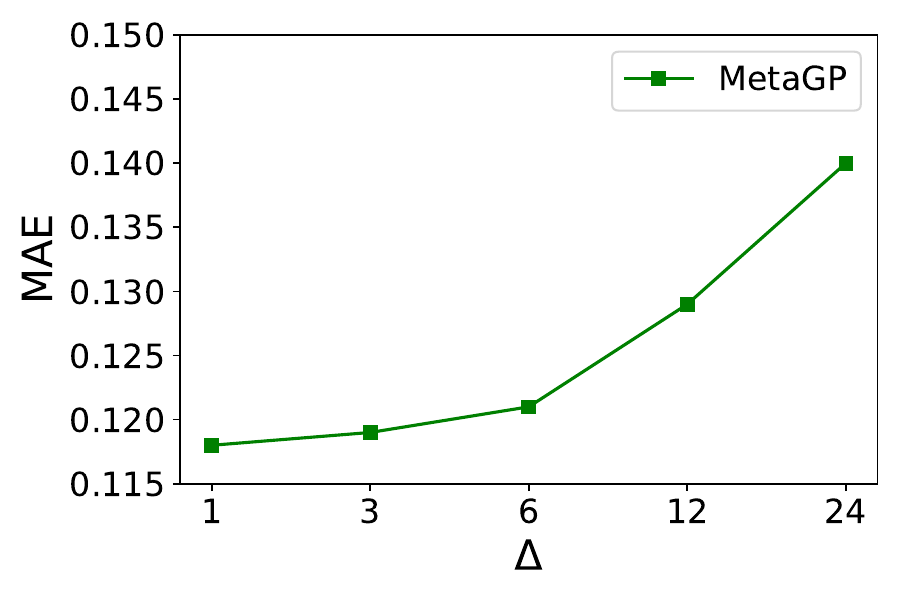}}
\subfigure[Training Time Vary with $\Delta$]{
\includegraphics[width=0.231\textwidth]{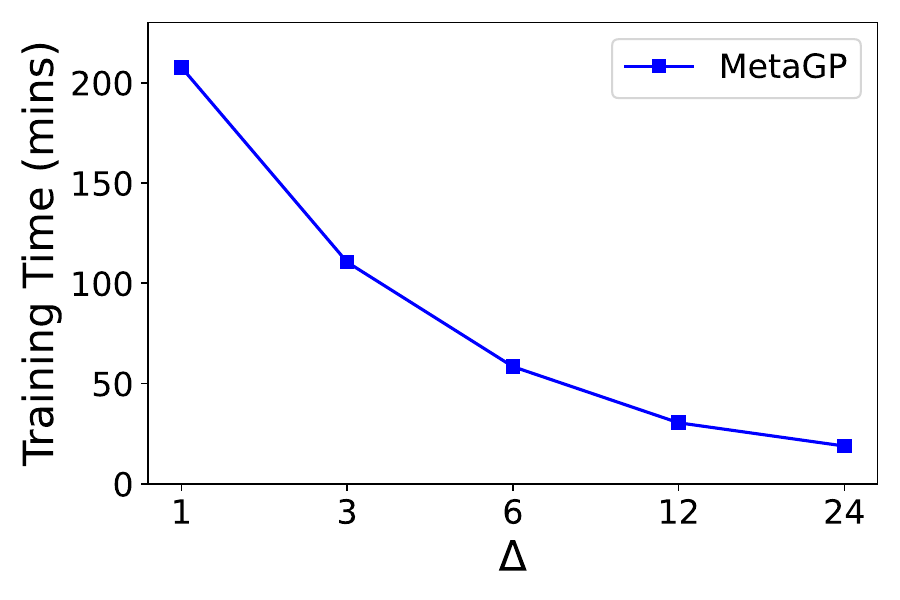}}
\subfigure[MAE Vary with $R$]{
\includegraphics[width=0.231\textwidth]{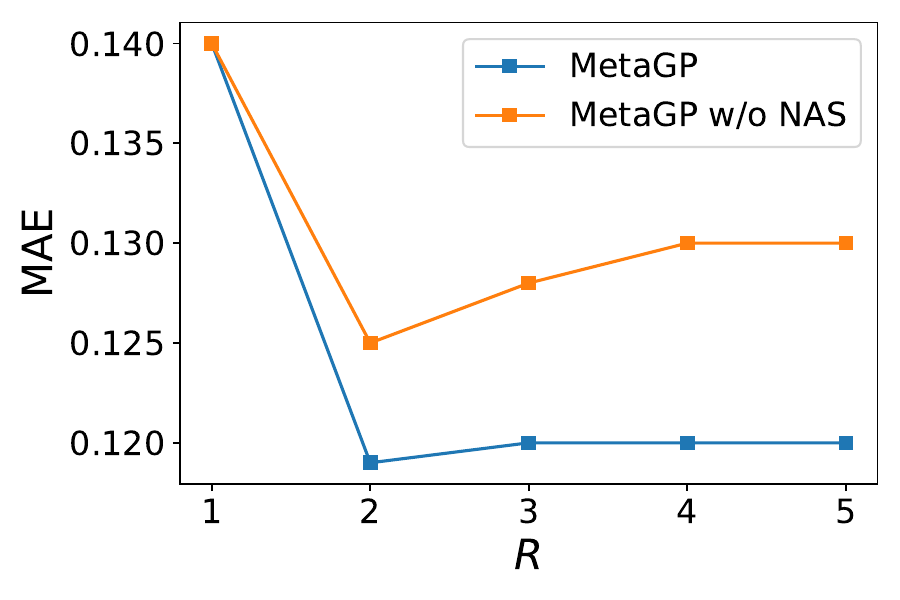}}
\subfigure[Training Time Vary with $R$]{
\includegraphics[width=0.231\textwidth]{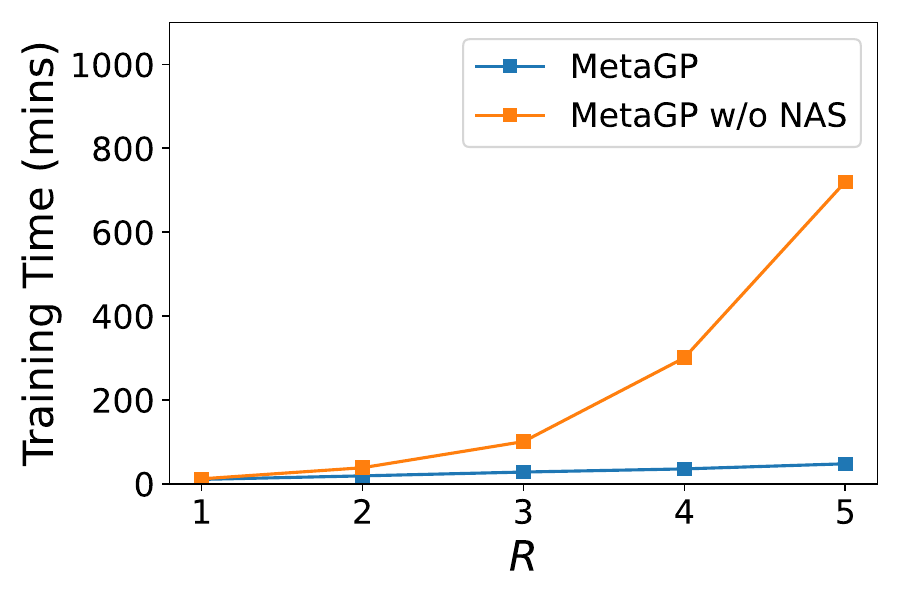}}
\caption{Hyperparameter Sensitivity.}
\label{hyper}
\end{figure*}

We proceed to consider the impact of changes in MetaGP's hyperparameters \(\Delta\) and \(R\) on its prediction accuracy and training time. All experiments are conducted on MC. Additionally, we use the full meta-training set and set the historical horizon \(L\) to 48 while keeping the forecasting horizon \(F\) at 12 to obtain more experimental results.

As shown in~\autoref{hyper}(a), the MAE of MetaGP increases with \(\Delta\). This is because the overlap and information density between input subsequences decreases, which decreases the accuracy. Moreover, as shown in~\autoref{hyper}(b), the training time of MetaGP decreases with a larger \(\Delta\), as using fewer input sequences accelerates training. We can thus use hyperparameter \(\Delta\) to balance between the computational resources used and the prediction accuracy.

Next, on the M4 dataset, we vary depth \(R\) to assess the sensitivity to \(R\). MetaGP w/o NAS employs a greedy strategy for meta-learning, and we use it as a baseline. Due to the high time complexity of MetaGP w/o NAS, we set \(\Delta\) to 24, as this reduces the training time.~\autoref{hyper}(c) shows that MetaGP and MetaGP w/o NAS achieve their highest prediction accuracy when \(R=2\) or \(R=3\). Although a larger \(R\) yields more high-order hybrid patterns for fitting the training data, this also increases the risk of overfitting. Finally, we report the training time of KAS and greedy kernel search.~\autoref{hyper}(d) shows that MetaGP has a lower training time than MetaGP w/o NAS. Therefore, MetaGP's KAS, which employs neural architecture search, not only avoids local optima and maintains a prediction accuracy that is at least as good as that of the greedy strategy, it also reduces the time complexity of the greedy strategy. We recommend setting \(R\) to 2 or 3 to maintain a short training time without jeopardizing prediction accuracy.

\subsection{Robustness Study}
To evaluate the robustness of MetaGP, we introduce outliers with a magnitude of five times the standard deviation into the Gravity-16 and MC datasets. The outlier ratios are set to 1\%, 3\%, and 5\%, representing the proportion of samples containing outliers relative to the total number of samples. We compare with meta-GRU-res, the most accurate method among the baselines.

As shown in~\autoref{robustness}, the accuracy of both methods drops as the outlier ratio increases. However, MetaGP consistently achieves the best accuracy. This indicates that although few-shot learning is affected by outliers, MetaGP maintains superior robustness.
\begin{table}[htbp]
\footnotesize
\centering
\caption{Robustness Study on Gravity-16 and MC.}
\label{robustness}
\setlength\tabcolsep{3pt}
\begin{tabular}{c|c|c|c|c}
\Xhline{1pt}
Dataset                     & Ratio        & 1                      & 3                      & 5                      \\ \Xhline{1pt}
\multirow{2}{*}{Gravity-16} & MetaGP       & \textbf{2.14 (0.88)}    & \textbf{2.32 (0.91)}    & \textbf{2.54 (0.93)}    \\
                            & meta-GRU-res & 3.05 (1.49)             & 3.34 (1.55)             & 3.59 (1.63)             \\ \hline
\multirow{2}{*}{MC}         & MetaGP       & \textbf{1.27 (0.15)e-1} & \textbf{1.30 (0.15)e-1} & \textbf{1.37 (0.17)e-1} \\
                            & meta-GRU-res & 1.43 (0.19)e-1          & 1.49 (0.20)e-1          & 1.61 (0.23)e-1             \\ \Xhline{1pt}
\end{tabular}
\end{table}

\subsection{Visualization}
Next, we visualize different meta-knowledge patterns and multivariate correlations captured by MetaGP.
\begin{figure*}[htbp]
\centering
\subfigure[Prediction vs. Ground Truth]{
\includegraphics[width=0.18\textwidth]{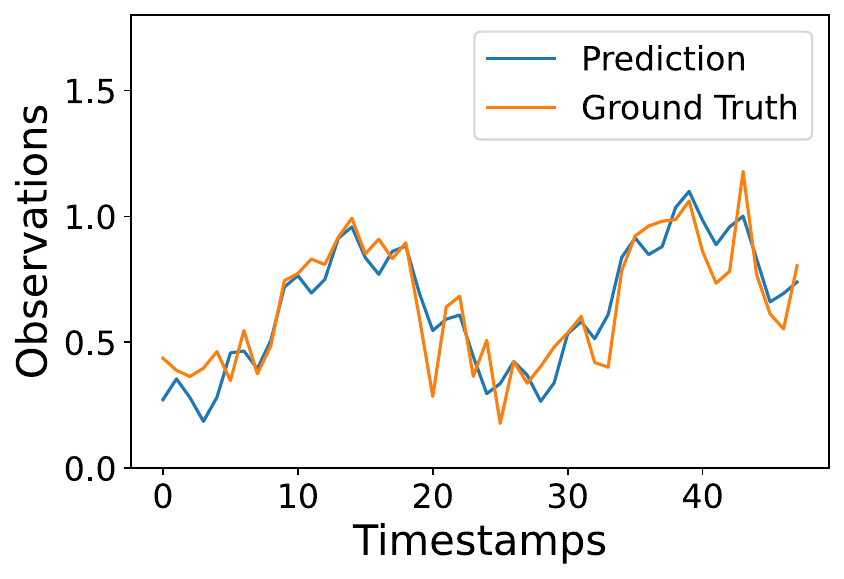}}
\subfigure[Long-term Seasonality Captured by the PER Kernel]{
\label{meta_patterns:b}
\includegraphics[width=0.18\textwidth]{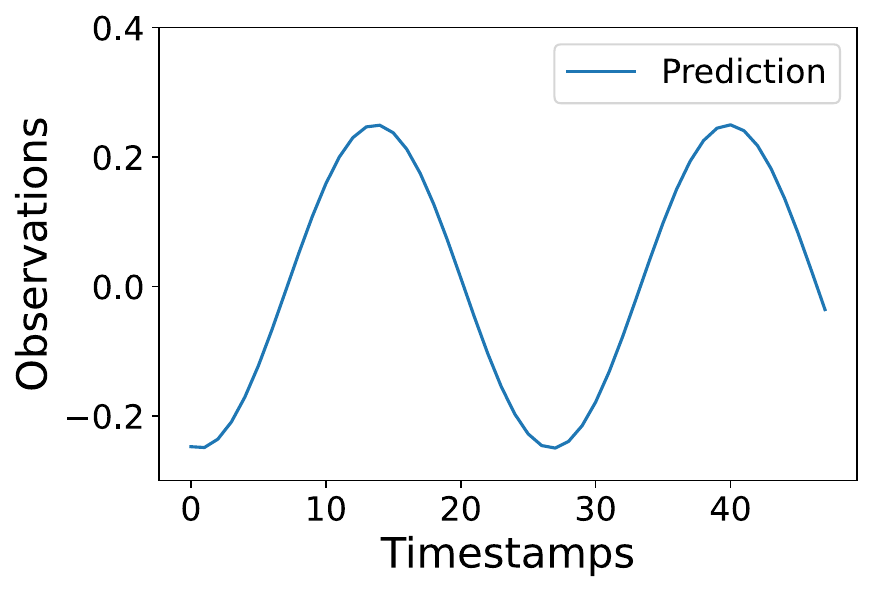}}
\subfigure[Trend Captured by the LIN Kernel]{
\includegraphics[width=0.18\textwidth]{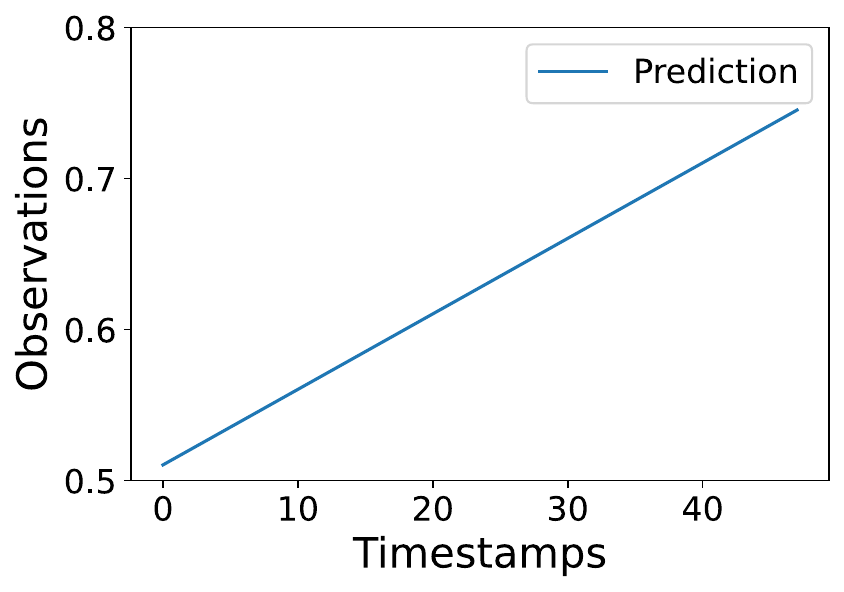}}
\subfigure[Mixed Short-term Dependency and Seasonality]{
\label{meta_patterns:d}
\includegraphics[width=0.18\textwidth]{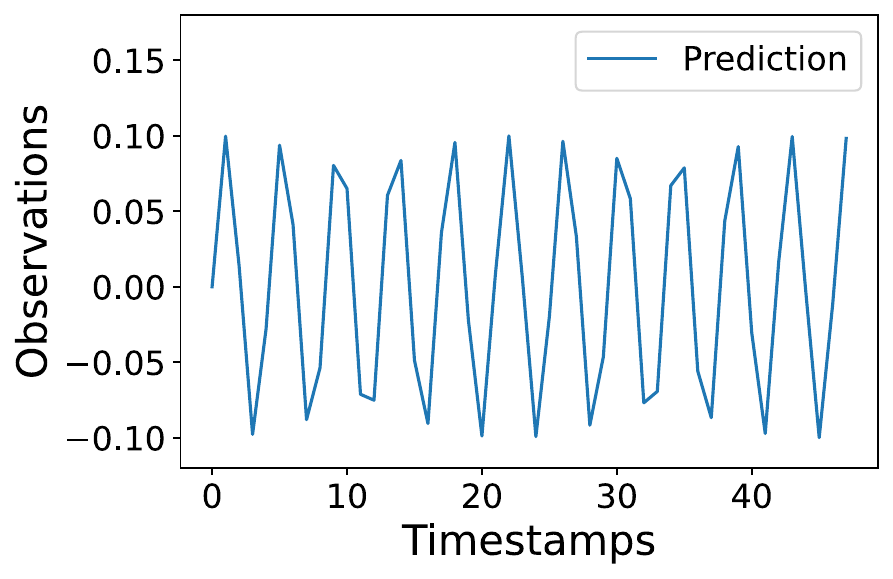}}
\subfigure[Mixed Trend and Long-term Seasonality]{
\includegraphics[width=0.18\textwidth]{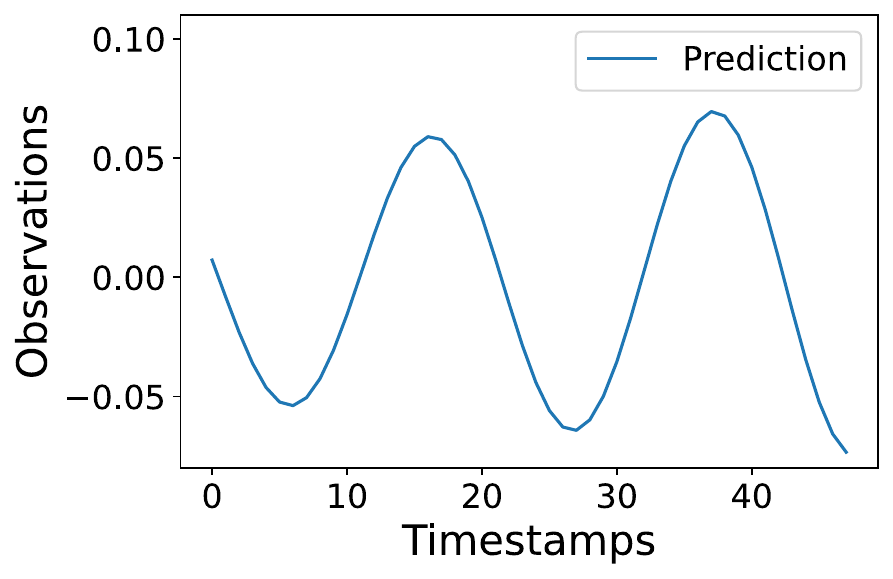}}
\caption{Visualization of Meta-knowledge Patterns.}
\label{meta_patterns}
\end{figure*}

\subsubsection{Visualization of Meta-knowledge Patterns} We extract a sample from MC that contains different patterns to assess the ability of KAS to model and capture different meta-knowledge patterns.

\autoref{meta_patterns}(a) shows MetaGP's predictions and the corresponding ground truth for the sample. MetaGP can fit the forecasting horizon well. This prediction is generated by the meta-knowledge learned by KAS. We rank the generated kernel functions by their weight and present the predictions produced by the top four functions; see Figures~\autoref{meta_patterns:b}--(e).

Figures~\autoref{meta_patterns:b}--(c) show two simple patterns: long-term seasonality and trend, modeled and captured by the PER and LIN kernels learned by KAS. Next, Figures~\autoref{meta_patterns:d}--(e) show two complex patterns: mixed short-term dependency and seasonality, and mixed trend and long-term seasonality, modeled and captured by the SE$\times$PER and LIN$\times$RQ kernels learned by KAS.

\subsubsection{Visualization of Correlations} There are two types of correlations between variables: one is the specific correlation between variables of two input sequences, as exemplified in~\autoref{fig:Interpretations}(a); the other is the generalized correlation that applies broadly to one type of data, as exemplified in~\autoref{fig:Interpretations}(b). 
\begin{figure}[htbp]
\centering
\subfigure[Multivariate Covariance Matrix ]{
\label{fig:subfig:type3} 
\includegraphics[width=0.231\textwidth]{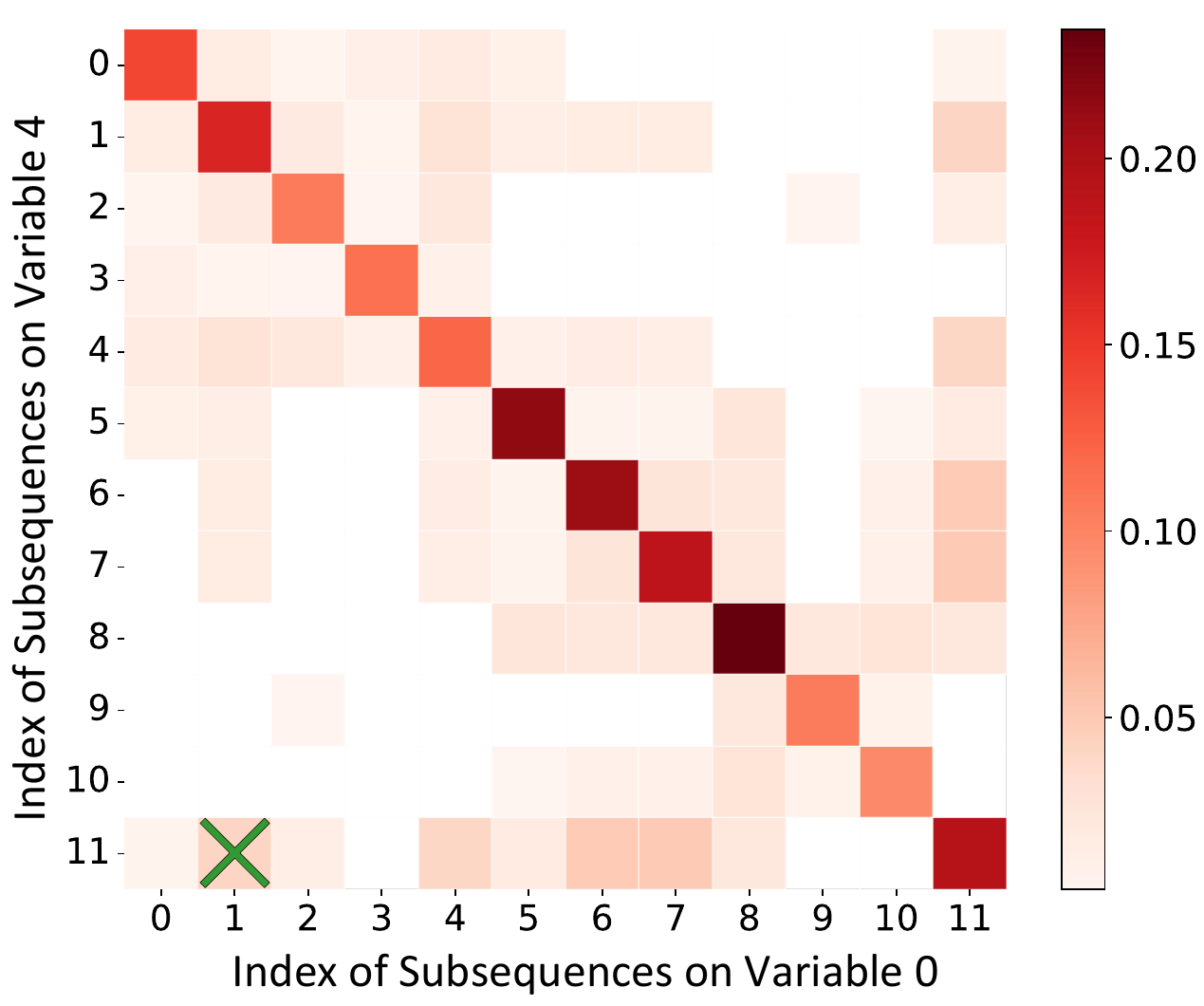}}
\subfigure[Cross-variable Weights]{
\label{fig:subfig:type4} 
\includegraphics[width=0.231\textwidth]{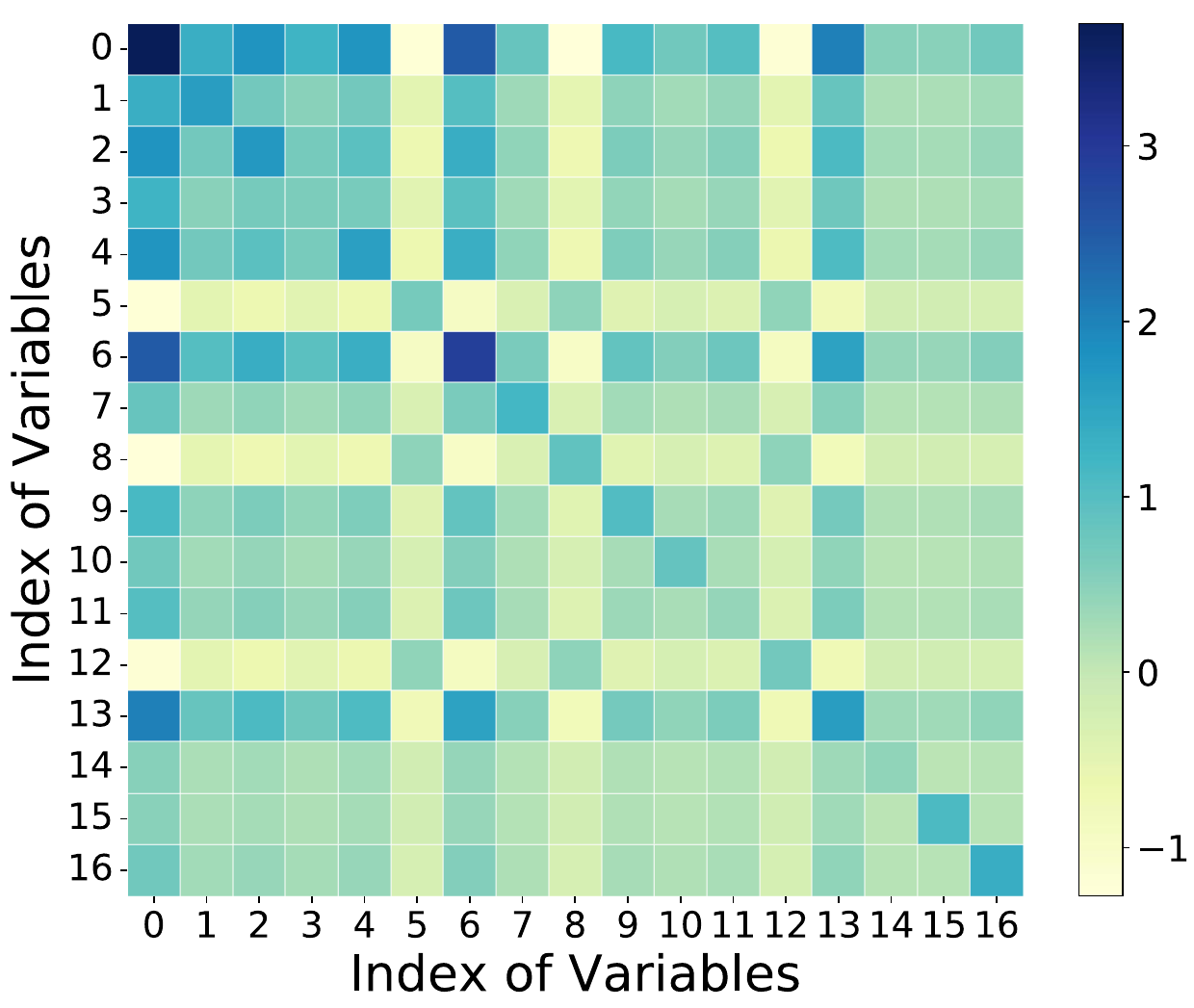}}
\caption{Visualization of Two Types of Correlations.}
\label{fig:Interpretations} 
\end{figure}

MetaGP utilizes a multivariate covariance matrix to capture the correlations between each two input sequences. An example matrix that captures the correlation between variables 0 and 4 is shown in~\autoref{fig:Interpretations}(a), where darker colors indicate higher positive correlations and lighter colors indicate higher negative correlations. We observe that the correlations between different input sequences can be captured. For example, input sequence 1 of variable 0 has a positive correlation with input sequence 11 of variable 4; see the cross in~\autoref{fig:Interpretations}(a). The values in the covariance matrix reflect the true correlations between specific samples and can be applied in scientific analyse. 

Then, we show cross-variable weights in~\autoref{fig:Interpretations}(b). The higher an absolute weight value is, the higher the importance of the corresponding variable. The clear differences between the cross-variable weights indicate that the dynamics between variables in a time series are highly complex, making it difficult to describe them with a single dynamic function. This underscores the necessity of this module. Moreover, cross-variable weights reflect a global and stable correlation between variables. This reflects the stable numerical relationship between physical quantities.

\section{CONCLUSION AND FUTURE WORK} 
\label{conclusion}
This study addresses the need for effective few-shot time series forecasting, especially given the limited sample sizes in fields like physics and biology. We present the meta-learning Gaussian process latent variable model, MetaGP, which effectively addresses the challenges of capturing long-term dependencies and explicitly modeling of diverse meta-knowledge. By employing a Gaussian process, MetaGP maintains strong correlations across distant observations, overcoming the limitations of existing sequential latent variable models. Additionally, MetaGP's Kernel Association Search (KAS) component enhances its ability to capture meta-knowledge and correlations within multivariate time series, improving both prediction accuracy and interpretability. Extensive experiments show that MetaGP is capable of state-of-the-art performance on simulated and real-world biological data, confirming its effectiveness at few-shot forecasting. The latent state architecture of MetaGP requires re-execution of the meta-learning process to discover meta-knowledge when transferring to different domains. However, as knowledge across related domains may share common structures, this re-execution may be unnecessary. In future research, we aim to improve the meta-learning by leveraging the generalization and knowledge summarization capabilities of large language models. This integration is expected to improve the cross-domain deployability of MetaGP while increasing both efficiency and accuracy. 


\vfill

\end{document}